\RequirePackage{rotating}

\documentclass[sigconf]{acmart}

\usepackage{pdflscape} %
\usepackage{ragged2e} %
\usepackage{colortbl} %

\usepackage{booktabs, multirow} %
\usepackage{soul}%
\usepackage{pifont} %
\usepackage{graphicx}
\usepackage{rotating,acmart-taps}

\newcommand*\rot{\rotatebox{90}}
\newcommand{\cmark}{\ding{51}}
\newcommand{\xmark}{\ding{55}}

\AtBeginDocument{%
  \providecommand\BibTeX{{%
    \normalfont B\kern-0.5em{\scshape i\kern-0.25em b}\kern-0.8em\TeX}}}

\copyrightyear{2023}
\acmYear{2023}
\setcopyright{rightsretained}
\acmConference[FAccT '23]{2023 ACM Conference on Fairness, Accountability, and Transparency}{June 12--15, 2023}{Chicago, IL, USA}
\acmBooktitle{2023 ACM Conference on Fairness, Accountability, and Transparency (FAccT '23), June 12--15, 2023, Chicago, IL, USA}
\acmDOI{10.1145/3593013.3593997}
\acmISBN{979-8-4007-0192-4/23/06}

\begin{document}

\title{Saliency Cards: A Framework to Characterize and Compare Saliency Methods}

\author{Angie Boggust}
\authornote{Both authors contributed equally to this research.}
\email{aboggust@csail.mit.edu}
\affiliation{%
  \institution{MIT CSAIL}
  \city{Cambridge}
  \state{Massachusetts}
  \country{USA}
}
\author{Harini Suresh}
\authornotemark[1]
\email{hsuresh@mit.edu}
\affiliation{%
  \institution{MIT CSAIL}
  \city{Cambridge}
  \state{Massachusetts}
  \country{USA}
}

\author{Hendrik Strobelt}
\affiliation{%
  \institution{IBM Research}
  \city{Cambridge}
  \state{Massachusetts}
  \country{USA}}

\author{John Guttag}
\affiliation{%
  \institution{MIT CSAIL}
  \city{Cambridge}
  \state{Massachusetts}
  \country{USA}
}

\author{Arvind Satyanarayan}
\affiliation{%
 \institution{MIT CSAIL}
  \city{Cambridge}
  \state{Massachusetts}
  \country{USA}}

\renewcommand{\shortauthors}{Boggust and Suresh, et al.}

\begin{abstract}
  Saliency methods are a common class of machine learning interpretability techniques that calculate how important each input feature is to a model's output. 
We find that, with the rapid pace of development, users struggle to stay informed of the strengths and limitations of new methods and, thus, choose methods for unprincipled reasons (e.g., popularity).
Moreover, despite a corresponding rise in evaluation metrics, existing approaches assume universal desiderata for saliency methods (e.g., faithfulness) that do not account for diverse user needs.
In response, we introduce \emph{saliency cards}: structured documentation of how saliency methods operate and their performance across a battery of evaluative metrics. 
Through a review of 25 saliency method papers and 33 method evaluations, we identify 10 attributes that users should account for when choosing a method.
We group these attributes into three categories that span the process of computing and interpreting saliency: \emph{methodology}, or how the saliency is calculated; \emph{sensitivity}, or the relationship between the saliency and the underlying model and data; and, \emph{perceptibility}, or how an end user ultimately interprets the result. 
By collating this information, saliency cards allow users to more holistically assess and compare the implications of different methods.
Through nine semi-structured interviews with users from various backgrounds, including researchers, radiologists, and computational biologists, we find that saliency cards provide a detailed vocabulary for discussing individual methods and allow for a more systematic selection of task-appropriate methods.
Moreover, with saliency cards, we are able to analyze the research landscape in a more structured fashion to identify opportunities for new methods and evaluation metrics for unmet user needs. 
\end{abstract}



\maketitle

\section{Introduction}

As machine learning (ML) systems are deployed in real-world contexts, stakeholder interviews~\citep{tonekaboni2019clinicians,bhatt2020explainable}, design best practices~\citep{amershi2019guidelines}, and legal frameworks~\citep{EUdataregulations2018} have underscored the need for explainability. 
Saliency methods\,---\,a class of explanation methods that identify input features important to an ML model's output\,---\,are frequently used to provide explanations.
Saliency methods have helped ML researchers evaluate new models~\citep{boggust2021shared, kulesza2015principles}, clinicians make AI-assisted patient care decisions~\citep{porumb2020precision}, and users deploy fair and generalizable models~\citep{ribeiro2016should}.
As the popularity of saliency methods has grown, the number and diversity of saliency methods have correspondingly increased~\citep{erhan2009visualizing, simonyan2013deep, sundararajan2017axiomatic, selvaraju2017grad, springenberg2014striving, ribeiro2016should, lundberg2017unified, fong2017interpretable, kapishnikov2019xrai, smilkov2017smoothgrad, petsiuk2018rise}.
However, since each saliency method operates differently according to its algorithmic goals, conflicts have arisen as multiple methods can produce varying explanations for the same model and input~\citep{krishna2022disagreement}.

Researchers have proposed metrics to evaluate the effectiveness of saliency methods~\citep{li2021experimental,ding2021evaluating,tomsett2020sanity,adebayo2018sanity, zhang2018top}.
While promising, these approaches assume universal desiderata all saliency methods must achieve to be worth considering.
Evaluations are often described as ``tests''~\cite{ding2021evaluating}, ``sanity checks''~\cite{adebayo2018sanity}, or ``axioms''~\citep{sundararajan2017axiomatic}, suggesting the existence of an \textit{ideal} saliency method that passes every possible evaluation. 
However, this framing overlooks that saliency methods are \textit{abstractions} of model behavior.
They cannot offer a complete or wholly accurate reflection of a model's behavior (akin to a printout of model weights), so saliency methods must decide what information to preserve and sacrifice. 
Critically, saliency method abstraction decisions are motivated by downstream human-centric goals, such as generalizability~\citep{carter2019made, ribeiro2016should, lundberg2017unified}, algorithmic simplicity~\citep{carter2019made}, or perceptibility~\citep{smilkov2017smoothgrad, kapishnikov2019xrai}. 
Given the rich diversity in end-user expertise and needs~\cite{krishna2022disagreement,suresh2021beyond}, it is unlikely that one set of abstraction decisions will support all users, contexts, and tasks. 
How, then, should end users characterize and compare saliency methods to choose the most suitable one for their particular application?

The lack of standardized documentation for saliency methods and evaluative metrics makes it challenging to determine the benefits and limitations of particular methods and identify differences between them.
Without resources equivalent to \textit{model cards}~\citep{mitchell2019model} or \textit{datasheets}~\citep{gebru2018datasheets}, users are left to reference a potential sequence of research papers, including the original saliency method and all subsequent evaluations.
Given the influx of saliency method research, this is a prohibitively time-consuming process and is especially out-of-reach for the broad class of users without a research background in machine learning, like clinicians, lawmakers, and engineers. 
Moreover, this piecemeal assembly of information considers each evaluative result in isolation, making it challenging to reason about whether desirable properties of a saliency method (e.g., input sensitivity~\citep{yeh2019on} and minimality~\citep{carter2019made}) may be in tension with one another.
As a result, and as we find through a series of interviews, users currently select saliency methods in unprincipled manners, such as choosing a method based on its popularity instead of a thorough understanding of its strengths and limitations.

In response, we introduce \textit{saliency cards}: a structured documentation of how a saliency method is designed, operates, and performs across evaluative metrics.
Reflecting the diversity of user needs, we identify ten attributes of saliency methods that users may wish to consider when choosing a particular approach. 
To facilitate comprehension, we group these attributes into three categories corresponding to different parts of the process of computing and interpreting saliency: \textit{methodology}, or how the saliency is calculated; \textit{sensitivity}, or relationships between the saliency and the model, input, or label; and \textit{perceptibility}, or how a user perceives the output saliency. 
By collating this information, saliency cards help surface a method's strengths and weaknesses more holistically than individual paper results. 
Moreover, by offering a standard structure and visual design, saliency cards allow users to more easily compare methods, and more carefully reason about tradeoffs that might otherwise have been unapparent in the method design.

To evaluate the usefulness of saliency cards, we conduct a semi-structured interview study with nine saliency method users from diverse backgrounds, including saliency method developers, ML researchers, radiologists, computational biologists, and consultants. 
We find that saliency cards help users systematically select a saliency method that meets their needs. 
While previously, users chose saliency methods based on popularity or familiarity, with saliency cards, users prioritized attributes based on their task requirements.
Using the visual format of saliency cards, users efficiently analyzed their prioritized attributes and weighed tradeoffs between saliency methods to uncover the method best suited to their task.
Saliency card attributes also provided a shared vocabulary to discuss saliency methods, enabling users to precisely communicate their preferences, regardless of their prior experience with machine learning.
Standardized documentation enabled side-by-side comparison, revealing ripe areas for future work, including saliency methods designed for a specific set of priorities, additional evaluation metrics for understudied attributes, and customized evaluations based on a user's data and models.

Saliency card templates and examples are available at:\\ \url{https://github.com/mitvis/saliency-cards}.
\section{Related Work}

Saliency methods (often referred to as feature attribution methods) are popular techniques for explaining a machine learning model's decision. 
Given an input, model, and target label, saliency methods compute a feature-wise importance score describing each feature's influence on the model's output for the target label. 
However, each saliency method computes these feature importances differently. 
Gradient-based methods, such as guided backpropagation~\citep{springenberg2014striving} and Grad-CAM~\citep{selvaraju2017grad}, compute importance using the model's gradients.  
Perturbation-based methods, such as SHAP~\citep{lundberg2017unified} and RISE~\citep{petsiuk2018rise}, measure importance by modifying input features and measuring the model's response. 
And path-based methods, such as integrated gradients~\citep{sundararajan2017axiomatic} and XRAI~\citep{kapishnikov2019xrai}, compute feature importances by comparing model outputs for the actual input to a meaningless input. 
While these granular categorizations~\citep{molnar2019} sort saliency methods based on algorithmic differences, they do not capture the complete set of considerations. 
Two gradient-based methods can operate differently, apply to separate tasks, and have distinct usage considerations. 
Saliency cards expand upon existing categorization criteria by documenting the saliency method's algorithm as well as other usage considerations, such as its hyperparameters and how to set them, dependence on model architectures, computational constraints, and sources of non-determinism.

As the popularity of saliency methods has grown, a related line of research has begun evaluating saliency methods' \textit{faithfulness}\,---\,i.e., their ability to accurately represent the model's decision-making process. 
These evaluations are varied and include measuring the impact of adversarial perturbations~\citep{ghorbani2019interpretation}, model randomization~\citep{adebayo2018sanity}, dataset shifts~\citep{kindermans2019reliability}, and input dropout~\citep{samek2016evaluating} on the saliency output. 
However, its common for saliency methods to pass some faithfulness tests while failing others~\citep{tomsett2020sanity}. 
These discrepancies reflect the fact that faithfulness is too broad of a goal for evaluating saliency methods.  
As abstractions of model behavior, saliency methods necessarily preserve and sacrifice information in service of other human-centered goals such as simplicity or perceptibility.  
Depending on the use case, a user may accept a method that performs poorly on an evaluation that is low-priority for their task.  
Saliency cards lend a structure to these existing evaluating methods by splitting the concept of faithfulness into granular attributes that can inform tradeoffs and usage decisions for specific tasks.
Saliency cards group evaluation methods that test similar concepts, such as a saliency method's response to label perturbations~\citep{adebayo2018sanity, yang2019benchmarking}, while drawing distinctions between those that test other factors, such as consistency across models~\citep{ding2021evaluating} or saliency localization~\citep{zhang2016top}.

Saliency cards are complementary to documentation standards for machine learning datasets~\citep{gebru2018datasheets, arnold2019factsheets, holland2020dataset, bender2018data, mcmillan2021reusable, hutchinson2021towards, diaz2022crowdworksheets, pushkarna2022data} and models~\citep{mitchell2019model, mcmillan2021reusable, crisan2022interactive, adkins2022prescriptive, seifert2019towards, shen2021value}.
These transparency artifact have been widely adopted by the ML community and led to increased trust and dissemination~\citep{mcmillan2021reusable, huggingfacemodelcard, googlemodelcard}.
However, there is no standardized procedure for releasing saliency methods.
Consequently, when selecting an appropriate saliency method, users must reference the original paper and subsequent evaluations to understand the algorithm, its advantages and limitations, and how to use it effectively.
This process is time-consuming for all users, but it is particularly prohibitive for users with little academic ML training, such as clinicians, lawmakers, and engineers.
Saliency cards address this gap by providing a documentation structure and surfacing useful considerations about saliency methods to a range of stakeholders.

\section{The Structure of Saliency Cards}
\begin{figure*}[t]
  \centering
  \includegraphics[width=\textwidth]{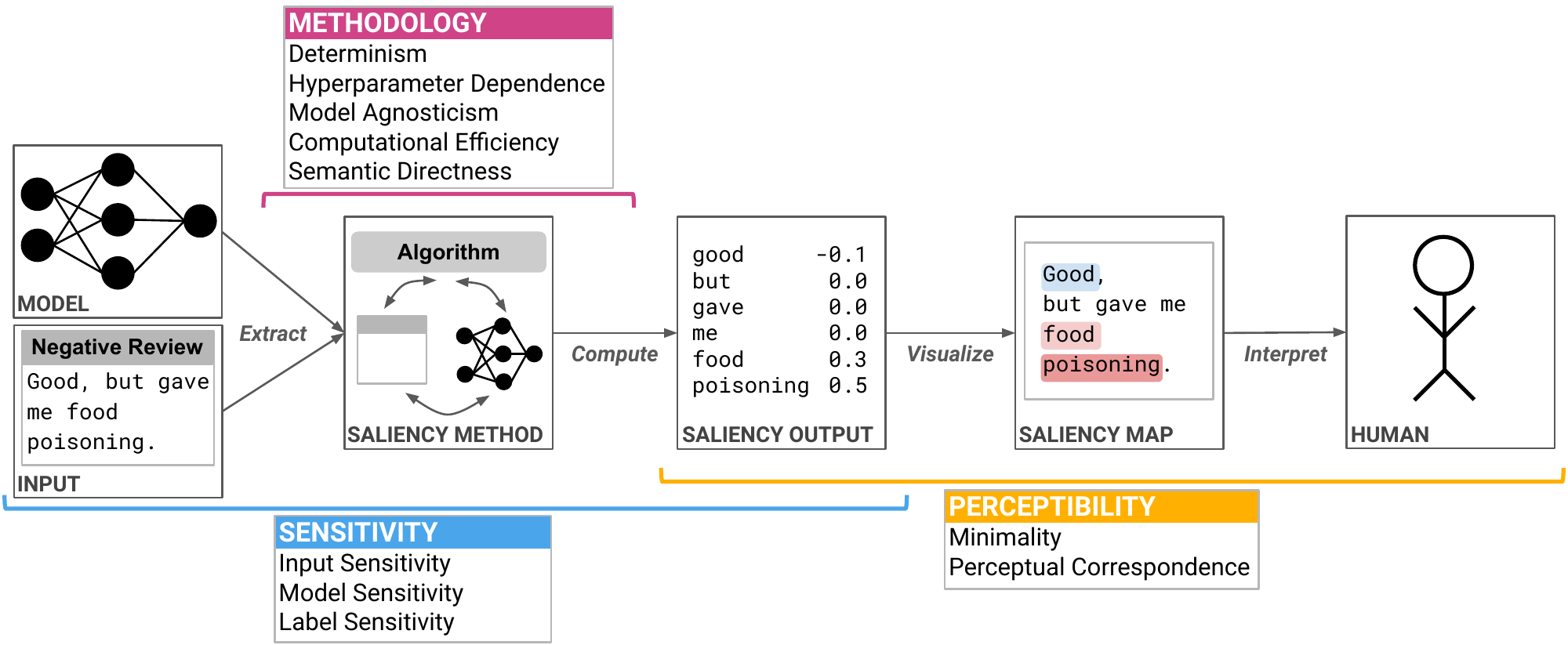}
  \caption{Saliency cards characterize saliency methods via ten user-centric attributes grouped into three categories corresponding to different phases of the interpretation process. \textit{Methodology} attributes describe how the saliency is computed, \textit{sensitivity} attributes express relationships between the saliency and its inputs, and \textit{perceptibility} attributes measure human perception of the saliency.}
  \label{fig:pipeline}
  \Description{A diagram shows where the categories are located within the saliency method pipeline. From left to right, the model and input feed into the saliency method which outputs the saliency output which is computed into a saliency map which is interpreted by a human. Methodology spans the saliency method. Sensitivity spans the model and input to the saliency output. Perceptibility spans the saliency output to the human.}
\end{figure*}

\begin{figure*}[t]
  \centering
  \includegraphics[width=\textwidth]{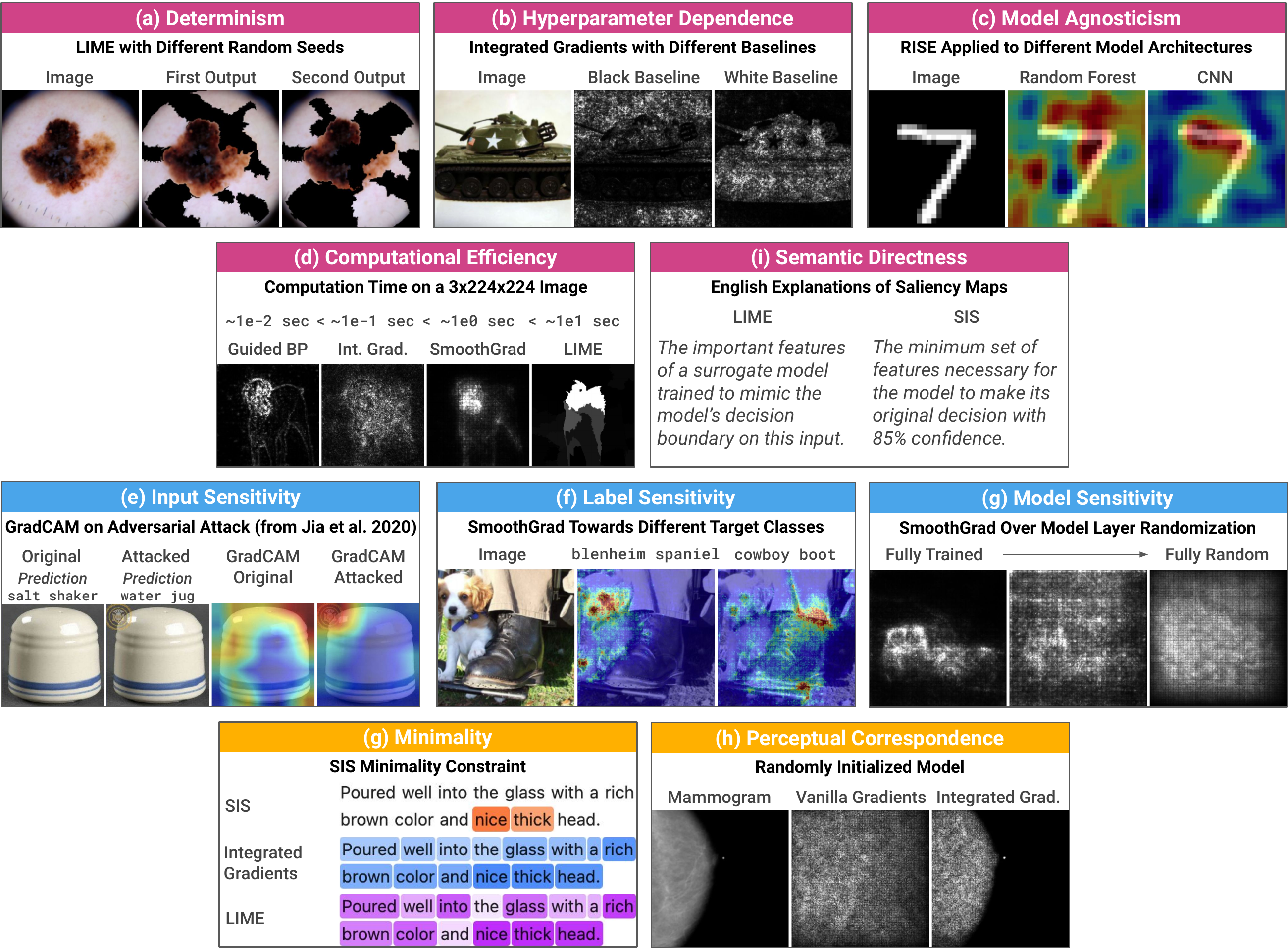}
  \caption{The ten saliency card attributes represent saliency method characteristics that can help users choose a task-appropriate saliency method. Each attribute is shown above alongside a descriptive example that communicates the principle. }
  \label{fig:attributes}
  \Description{Examples for each of the ten saliency card attributes. Each example shows the title of the attribute and an illustrative example from computer vision tasks.}
\end{figure*}

Saliency cards summarize information about a saliency method, including its developers, design goals, input/model/user assumptions, dependencies, usage considerations, benefits and limitations, and performance across various evaluations. 
The cards are structured as a series of attributes, grouped into three categories.
We derived these attributes and categories through reviewing literature on saliency method algorithms, evaluation metrics used to assess them, and commonly-cited desiderata for model explanations.  
Our approach entailed iteratively 1) finding commonalities across methods, stated desiderata, and evaluative metrics, 2) distilling attributes that captured these commonalities, and 3) applying these attributes to compare and understand a broad set of saliency methods. 
This process yielded ten attributes grouped into three higher-level categories: \textit{methodology}, or how the method operates; \textit{sensitivity}, or how the method responds to changes in the model or data; and \textit{perceptibility}, or how a human interprets the result of the method (Fig.~\ref{fig:pipeline}).
For each attribute, we provide a visual example (Fig.~\ref{fig:attributes}) and real-world application from our user study (Sec.~\ref{sec:user-study}).

\subsection{Methodology}
The methodology section of a saliency card summarizes the method's algorithm, including references to demos, papers, and/or implementations.  
In addition, this section includes information about five attributes related to how the saliency method operates: determinism, hyperparameter dependence, model agnosticism, computational efficiency, and semantic directness.
By documenting these details, the methodology section provides an informative summary that helps users understand if a saliency method applies to their task. 

\subsubsection{Determinism}
Determinism measures if a saliency method will always produce the same saliency map given a particular input, label, and model. 
Some saliency techniques, like LIME~\citep{ribeiro2016should} and SHAP~\citep{lundberg2017unified}, are non-deterministic, so running them multiple times can produce significantly different results. 
Non-determinism can be introduced in the algorithm's definition; for instance, by computing the many random masks used by RISE~\citep{petsiuk2018rise}.
It can also result from stochastic hyperparameters, like using random noise as a baseline value for integrated gradients~\citep{sundararajan2017axiomatic}.

Understanding how deterministic a saliency method is can impact if and how users apply the method. 
For example, in a clinical diagnosis task, a non-deterministic method could result in saliency maps highlighting slightly different portions of the radiograph.
The radiologist we interviewed (\texttt{U9}) worried that these variations may have significant consequences given that small areas of the image can be integral to the diagnosis.
Looking only at a single saliency map could skew a radiologist's judgment, while interpreting multiple maps together may be too time-consuming for the task.  
Thus, a user might choose to prioritize a deterministic saliency method in this setting. 
On the other hand, non-determinism can provide helpful additional context by surfacing multiple reasons for a model's decision. 
Just as humans can provide multiple correct justifications for their decisions, models likely have multiple feature sets sufficient to make correct and confident decisions. 
For example, in an image classification task, a model may correctly identify the object in the image using independent subsets of the object. 
A model developer we interviewed, \texttt{U5}, uses saliency to uncover spurious correlations between uninformative inputs and correct model outputs, so they were interested in leveraging non-determinism to surface all possible model justifications.

\subsubsection{Hyperparameter Dependence}
Hyperparameter dependence measures a saliency method's sensitivity to user-specified parameters. 
Some methods, like vanilla gradients~\citep{simonyan2013deep, erhan2009visualizing}, do not have hyperparameter settings or require user intervention. 
However, other methods, like integrated gradients~\citep{sundararajan2017axiomatic}, require users to set consequential hyperparameters whose ideal values vary drastically depending on the task.

By documenting a method's hyperparameter dependence, saliency cards inform users of consequential parameters and how to set them appropriately. 
Using default parameter values can be misleading if users do not have sufficient resources or expertise to devote to hyperparameter tuning.
Similarly, confusing results can arise if the hyperparameters were chosen based on a particular dataset but deployed in a setting with significant distribution shift. 
In situations like this, it makes sense for users to prioritize methods with low hyperparameter dependence. 
The radiologist (\texttt{U9}) prioritized hyperparameter dependence because they worried an incorrectly set parameter could have life-or-death consequences. 
Medical data often differs from the research datasets parameters are tuned on, so they worried a software vendor might not set consequential parameters appropriately.
For example, integrated gradients~\citep{sundararajan2017axiomatic} computes feature importance by interpolating between a ``baseline'' parameter and the actual input.  
A common practice is to use a baseline of all zeroes; however, using a zero baseline in x-ray images can be misleading and potentially harmful. 
In x-rays, black pixels convey meaning, such as indicating a bone fracture. 
If a software vendor uses the default zero baseline (black), 
integrated gradients will indicate that the fracture pixels are unimportant. 
The saliency card for integrated gradients would describe this hyperparameter dependence, alerting users to choose an appropriate baseline value or select a saliency method with less hyperparameter dependence.
On the other hand, when researchers have dedicated appropriate time and resources to hyperparameter tuning, it could be preferable to use a method dependent on hyperparameters, like SmoothGrad~\citep{smilkov2017smoothgrad}, because it has other desired attributes, like minimality (Sec.~\ref{sec:minimality}).

\subsubsection{Model Agnosticism}
Model agnosticism measures how much access to the model a saliency method requires. 
Model agnostic methods, such as SHAP~\citep{lundberg2017unified}, treat the underlying model as a black box, relying only on its input and output. 
On the other hand, model-dependent methods, like Grad-CAM~\citep{selvaraju2017grad}, require access to model internals. 
Model-dependent methods may have specific requirements, such as differentiability (e.g., gradient-based methods) or a specific model architecture (e.g., Grad-CAM requires a CNN~\citep{selvaraju2017grad}).

By documenting a saliency method's model agnosticism, saliency cards help users identify whether a particular method might be incompatible with their use case. 
For instance, computational biologists, like \texttt{U8}, use proprietary machine learning models hosted through web-based GUIs, where users upload their input data and the model returns its predictions.  
In this setting, it is impossible to use a saliency method that requires access to model internals. 
Users, including \texttt{U2}, might also need a model agnostic saliency method for use cases that involve comparing saliency maps across different kinds of models.  
On the other hand, model agnosticism is not priority for users, like \texttt{U1}, who only consider specific model architectures.
For these use cases, a user might choose a model-dependent method, like Grad-CAM~\citep{selvaraju2017grad}, to prioritize other attributes essential to their task, like increased model sensitivity (Sec.~\ref{sec:model-sensitivity}).

\begin{figure*}[t]
  \centering
  \includegraphics[width=\textwidth]{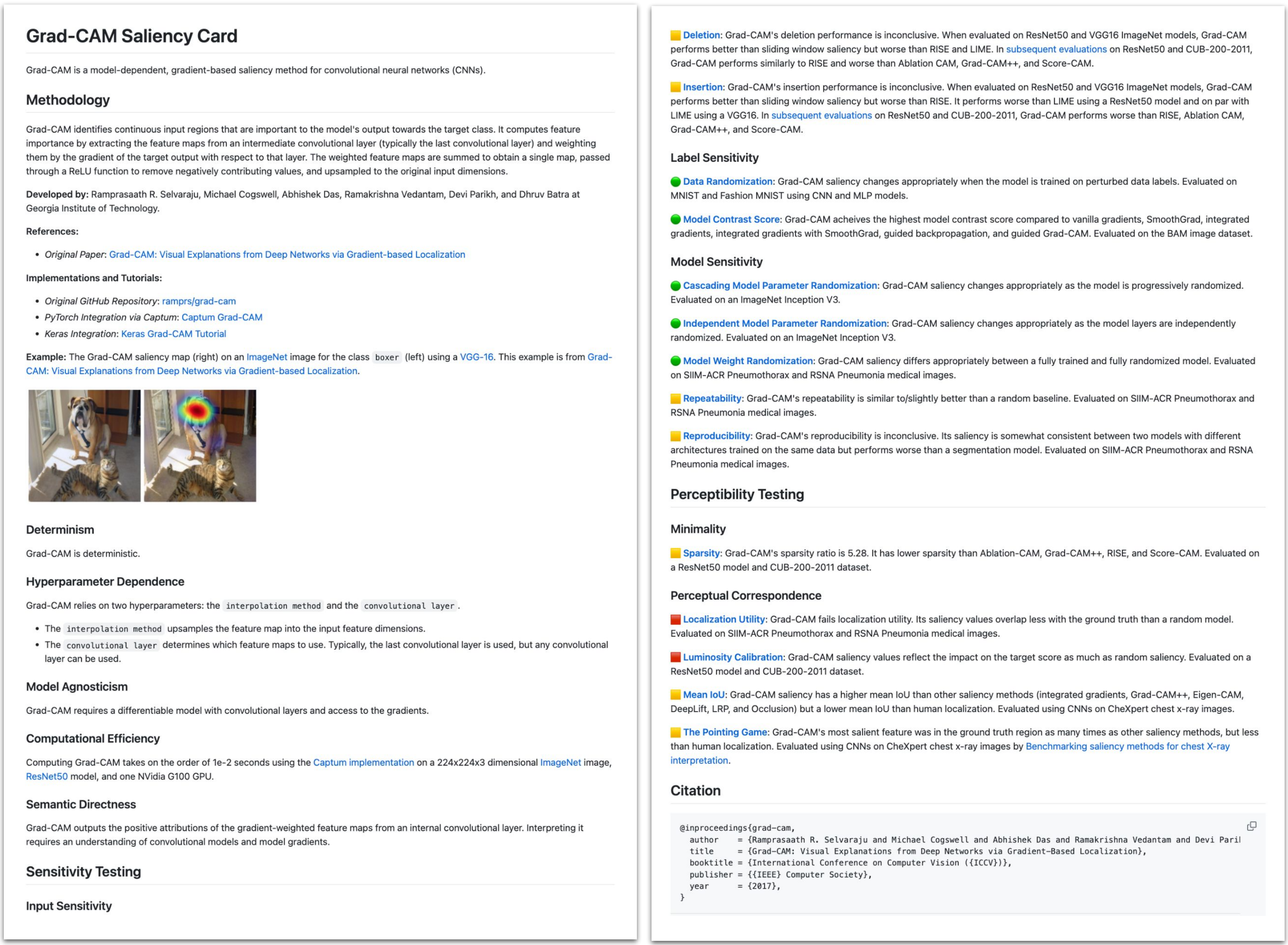}
  \caption{Saliency cards begin with a summary of the saliency method's algorithm, an example output, and references. Next, the card describes the saliency method's methodology, including its determinism, hyperparameter dependence, model agnosticism, computational efficiency, and semantic directness. Finally, the saliency card summarizes the method's performance on sensitivity (input, label, and model sensitivity) and perceptibility (minimality and perceptual correspondence) tests.}
  \label{fig:saliency-card}
  \Description{Side by side images show pages 1 and 2 of the Grad-CAM saliency card. The left page shows the methodology descriptions. The right page shows the sensitivity and perceptibility tests.}
\end{figure*}

\subsubsection{Computational Efficiency}
Computational efficiency measures how computationally intensive it is to produce the saliency map. 
Different methods vary widely in their computational efficiency. 
For example, perturbation-based methods, like meaningful perturbations~\citep{fong2017interpretable}, are often more computationally intensive than simple gradient-based methods, like guided backpropagation~\citep{springenberg2014striving}. 

By documenting computational efficiency, saliency cards help users determine whether running a particular saliency method is feasible in their setting.
Several details about the task\,---\,including the number of saliency maps to compute, the number of models to compare, the amount of available computational resources, the size of the input data\,---\,all play a role in whether or not to prioritize computational efficiency.
For example, some users, like \citet{global2019pfau} and \texttt{U1}, use saliency maps to compute aggregate statistics about their model's behavior across an entire dataset.
Given the size of existing machine learning datasets, this task could require computing hundreds of thousands of saliency maps to analyze a single model.  
In this setting, users may need to prioritize computational efficiency over other attributes.

\subsubsection{Semantic Directness}
Saliency methods abstract different aspects of model behavior, and semantic directness represents the complexity of this abstraction. 
For example, the saliency map computed by SIS represents a minimal set of pixels necessary for a confident and correct prediction~\citep{carter2019made}. 
Meanwhile, LIME's saliency map represents the learned coefficients of a local surrogate model trained to mimic the original model's decision boundary~\citep{ribeiro2016should}.

Semantically direct saliency methods do not require understanding complex algorithmic mechanisms such as surrogate models (e.g., as in LIME~\citep{ribeiro2016should}) or accumulated gradients (e.g., as in integrated gradients~\citep{sundararajan2017axiomatic}).  
As a result, their outputs might be more intuitive to users without formal ML expertise. 
Documenting semantic directness in a saliency card can help users prioritize it for tasks where the saliency maps will be interpreted by people with varying backgrounds.
For example, in our interviews, users who worked in mixed-experience teams, like \texttt{U8}'s biologist coworkers and \texttt{U7}'s business clients, prioritized semantic directness to help them efficiently and effectively communicate their results.
However, in cases where the interpreter can understand the saliency method or is comfortable not understanding the algorithm, semantic directness may not be a priority. 
We might choose a saliency method that is not semantically direct, like SHAP~\citep{lundberg2017unified} (which defines feature importances as game theoretic Shapley values), because it improves other attributes, like perceptual correspondence (Sec.~\ref{sec:perceptual-correspondence}).

Semantic directness is a methodological attribute because it describes the complexity of the saliency method's algorithm. 
However, it is also related to perceptibility (Sec.~\ref{sec:perceptibility}) because it informs what types of users will use a saliency method and how they will interpret its results. 
We choose to delineate the methodology and perceptibility attributes based on their testability\,---\, i.e., which attributes have been quantified in the literature. 
Currently, semantic directness is purely descriptive, and no saliency evaluations test for it. 
As evaluations are developed (Sec.~\ref{sec:user-study-future-work}), we may discover that semantic directness is testable and part of perceptibility or confirm it is part of methodology because it is unique to each user.

\subsection{Sensitivity Testing}
\label{sec:sensitivity}
The sensitivity section of a saliency card details whether a saliency method changes in proportion to meaningful changes in the model and data. 
While prior work has treated sensitivity as an overarching goal of all saliency methods, saliency cards break sensitivity down into three independent attributes: input sensitivity, label sensitivity, and model sensitivity. 
Saliency methods often perform differently across each of these attributes. 
By disentangling them, saliency cards let users prioritize and make trade offs based on their needs.

Unlike the methodology section, which contains descriptive information about each attribute, the sensitivity section enumerates experimental evaluations for each attribute, documenting the results and linking to the original evaluation.
As a result, saliency cards provide a glanceable representation of complex evaluations, helping users of all backgrounds evaluate a method's sensitivity.

\subsubsection{Input Sensitivity}
Input sensitivity measures if a saliency method accurately reflects the model's sensitivity to transformations in the input space. 
If an input transformation meaningfully affects the model's output (e.g., an adversarial attack), the saliency method should assign importance to the transformed features. 
Otherwise, if an input transformation does not change the model's output (e.g., a noise-based perturbation), then the saliency method should not assign additional importance to the modified features. 

Documenting a saliency method's input sensitivity is essential for tasks that use saliency to understand the impact of input changes, such as studying a model's robustness to adversarial attacks~\citep{jia2020adv}, or its reliance on sensitive input features (e.g., race or age)~\citep{balagopalan2022road}.
For instance, \texttt{U3}, a fairness researcher, tests models for \textit{counterfactual fairness}~\citep{kusner2017counterfactual} by confirming the model's decision on an input is the same even if its sensitive attributes are inverted.
Without input sensitivity, a saliency method might indicate that a sensitive feature is unimportant even if the model's decision is counterfactually unfair.
As a result, the method could mislead a user to trust and deploy a discriminatory model.
However, in other cases, users might choose to trade off input sensitivity for higher priority attributes.
For example, we interviewed ML researchers (\texttt{U5}, \texttt{U6}) who only work in controlled settings that are less dependent on input changes, including analysis of in-distribution data or comparison of different models on the same data.

Saliency cards contain the results of numerous existing input sensitivity tests~\citep{samek2016evaluating, alvares2018towards, ding2021evaluating, chattopadhyay2018grad, jung2021towards, gomez2022metrics, ancona2017unified, petsiuk2018rise, yeh2019on, sundararajan2017axiomatic, kindermans2019reliability, hooker2019benchmark, alvarez2018on, carter2019made}. 
Several of these tests measure whether the model's output changes significantly in response to perturbations of salient features and little in response to changes in non-salient features.
Extensions of them evaluate if the salient input features are sufficient to train a new, equally performant model~\citep{hooker2019benchmark, carter2019made}.
Other metrics test input sensitivity by measuring the saliency map's response to uninformative transformations of the input dataset~\citep{kindermans2019reliability} or adding small amounts of noise to the inputs~\citep{yeh2019on, alvarez2018on}.
Saliency cards should report the results of various input sensitivity tests to communicate a comprehesive overview of a saliency method's input sensitivity.

\subsubsection{Label Sensitivity}
Saliency maps are computed with respect to a particular target label.
Label sensitivity measures the saliency method's response to changes to that target label. 

Documenting label sensitivity is important for tasks that evaluate model behavior by changing the target label.
For example, in computational immunology, users like \texttt{U8} train multi-task models that take in an antibody and predict how well it binds to several different target molecules~\citep{liu2020antibody}. 
In this setting, \texttt{U8} wants to compare which antibody features are important to the model's prediction and how they differ across multiple targets.
To do so, they compare saliency maps on the same antibody with respect to different target labels. 
Without label sensitivity, the saliency maps may inaccurately reflect the difference in model reasoning for different target molecules. 
Since antibody binding is not fully understood, label insensitive saliency methods could mislead users about which features are biologically important.

Saliency cards include the results of label sensitivity tests. 
Existing label sensitivity evaluations swap labels in controlled ways and measure changes in the saliency map.
Tests include measuring how the saliency maps change in response to label randomization~\citep{adebayo2018sanity} and when switching from fine-grained to coarse-grained labels~\citep{yang2019benchmarking}.

\subsubsection{Model Sensitivity} 
\label{sec:model-sensitivity}
Model sensitivity measures if the output of a saliency method is sensitive to 
meaningful changes to the model parameters. 
If we modify the model significantly (e.g., by randomizing its weights), then the output of a model sensitive saliency method will reflect that change.

Model sensitivity is crucial for tasks that compare models.
For example, we interviewed an ML researcher (\texttt{U2}) who evaluates their training procedure by comparing saliency maps across models from different epochs.
They prioritized model sensitive methods to accurately reflect meaningful model changes. 
However, users may trade off model sensitivity for other desirable attributes in tasks where the model is not changing.
For instance, if a user is focused on the impact of modifying input features, they might trade off model sensitivity to get a highly input sensitive method.

Saliency cards document a method's model sensitivity using existing model sensitivity tests. 
Testing for model sensitivity involves evaluating how the output of a saliency method changes in response to known similarities or differences between models. 
Some methods test the consistency of saliency maps between similar models~\citep{ding2021evaluating, arun2020assessing, sundararajan2017axiomatic}, while others confirm that saliency maps sufficiently change due to model randomization~\citep{adebayo2018sanity} or combination~\citep{sundararajan2017axiomatic}.

\subsection{Perceptibility Testing}
\label{sec:perceptibility}
The perceptibility section of a saliency card describes attributes of a saliency method related to human perception of the saliency map. 
Perceptibility is split into two attributes: minimality and perceptual correspondence. 
Minimality captures the idea that a saliency map should highlight a minimal set of features, which can be an important consideration for users who visually analyze saliency maps. 
Perceptual correspondence measures if the interpreted signal in the saliency map reflects the features' importance and does not introduce misleading visual artifacts. 
As with the sensitivity attributes, saliency cards summarize the results of perceptibility tests.

\subsubsection{Minimality}
\label{sec:minimality}
Minimality measures how many unnecessary features are given a significant value in the saliency map. 
Methods, such as vanilla gradients~\citep{simonyan2013deep}, that attribute importance to many input features can produce noisy results that are difficult to interpret. 
On the other hand, methods like XRAI~\citep{kapishnikov2019xrai} incorporate minimality into their algorithms by attributing importance to higher-level feature regions instead of many individual features.

Documenting a method's minimality can alert users to the amount of noise they can expect in the saliency map. 
For example, minimality is particularly important when interpreting complex high-dimensional data, such as the long amino acid sequences our computational biologist interviewee (\texttt{U8}) uses.
In this case, it might be prohibitively difficult or time-consuming to interpret hundreds of important features, and a noisy saliency map could risk obscuring underlying signal. 
However, some users may prefer a less minimal saliency method depending on their task. 
For instance, some bioinformatics tasks operate on input sequences containing only nine amino acids~\citep{andreatta2016gapped}. 
In this setting, users might actually prefer a less minimal method that reveals every amino acid that influences the model's prediction, since it provides a complete picture of important features and is relatively easy to analyze.  

Saliency cards report a saliency method's performance on minimality metrics. 
Some minimality metrics compare the maximum and mean saliency values~\citep{gomez2022metrics}. 
The higher the ratio, the more minimal the saliency map is, since it is focused on only a few input features. 
Minimality can also be evaluated by testing if any salient feature can be removed without the model's confidence dropping below a chosen threshold~\citep{carter2019made}. 
For methods that are not inherently minimal, applying SmoothGrad~\citep{smilkov2017smoothgrad} can increase minimality by reducing the noise present in the saliency map; however, it may also impact other attributes of the original method.

\subsubsection{Perceptual Correspondence}
\label{sec:perceptual-correspondence}
Perceptual correspondence measures if the perceived signal in the saliency map accurately reflects the feature importance. 
For example, saliency maps with high perceptual correspondence should not contain visual artifacts that lead users to incorrectly infer signal.

Saliency cards document perceptual correspondence because it is crucial for high-risk settings where a misleading signal could provide an unwarranted justification for decisions or lead users down incorrect paths. 
However, perceptual correspondence may be less critical if the saliency method is used for large-scale analyses of model behavior that will aggregate one-off artifacts (i.e., when such artifacts occur arbitrarily and are washed out by averaging). 
For instance, users, like \texttt{U1}, who aggregate metrics on saliency maps across an entire dataset may be willing to prioritize computational efficiency over perceptual correspondence.

Saliency cards report the results of perceptual correspondence tests.
The most direct perceptual correspondence metric measures if a feature's visualized luminosity is calibrated to its importance~\citep{gomez2022metrics}.
Other perceptual correspondence evaluations compare salient features to human-defined ground truth features~\citep{zhang2016top,arun2020assessing, ding2021evaluating}. 
However, these metrics should only be used to evaluate saliency methods in test settings where the model is known to rely on ground truth features. 
Even high-performing models have been shown to rely on spurious features~\citep{carter2021overinterpretation, xiao2021noise}.
In those cases, a perceptually correspondent saliency method would correctly highlight features outside of the ground truth but get low perceptual correspondence scores.

\section{Evaluative User Studies}
\label{sec:user-study}
\begin{figure*}[ht]
  \centering
  \includegraphics[width=\textwidth]{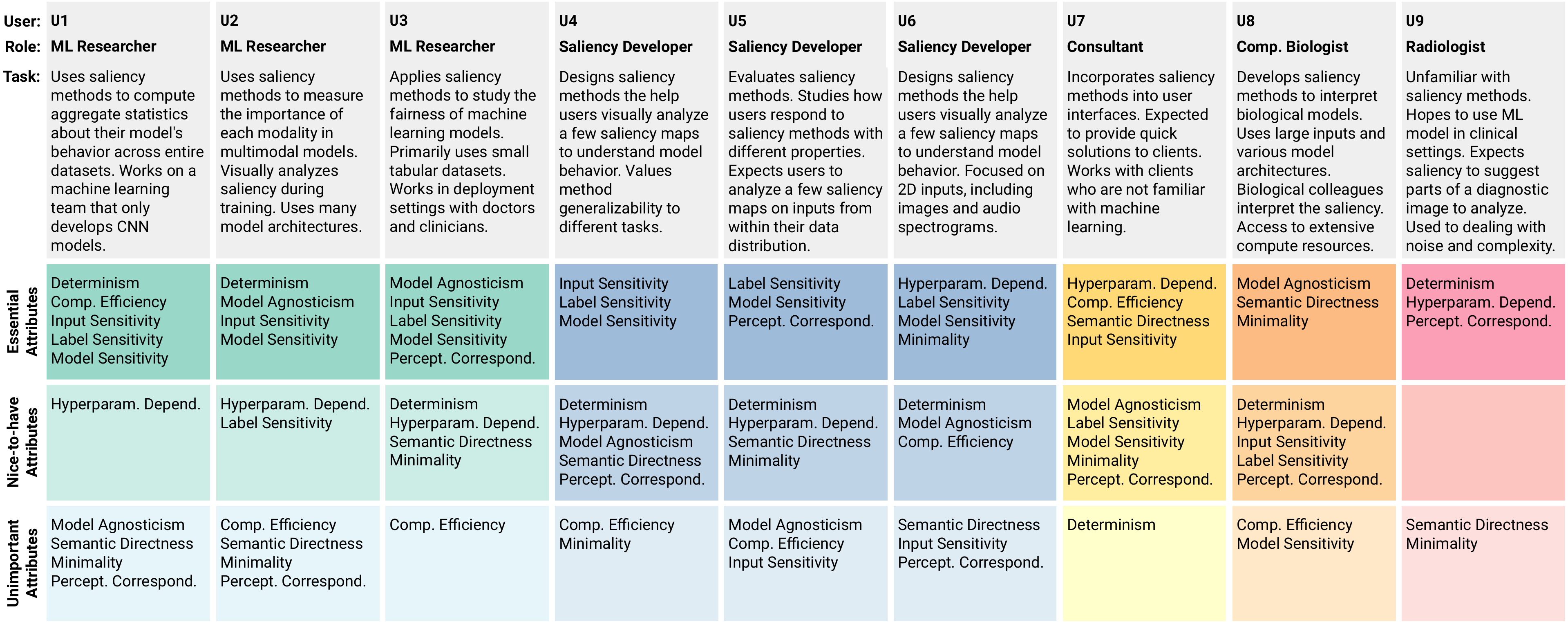}
  \caption{We evaluate saliency cards through semi-structured interviews with researchers, scientists, and domain experts. Each user prioritized saliency method attributes differently based on the needs of their tasks. Saliency cards helped users select task-appropriate methods, communicate about saliency methods, and hypothesize new areas for future work.}
  \label{fig:user-study}
  \Description{Nine sections describe each of the nine participants in the user study. There are three machine learning researchers, three saliency developers, one consultant, one computational biologist, and one radiologist.}
\end{figure*}

Through nine semi-structured user interviews, we evaluate saliency cards to understand how they can help users understand, compare, and select methods appropriate for their tasks. 
We recruited participants from within our professional network (4/9) and through an open call in our organizations and on Twitter (5/9). 
Participants came from diverse backgrounds (including research, radiology, computational biology, and consultancy) and had varying levels of machine learning expertise and familiarity with saliency methods.
Fig.~\ref{fig:user-study} illustrates user demographics, describes their saliency method use cases, and summarizes the results of our interview studies.

Eight participants (\texttt{U1}--\texttt{U8}) had used saliency methods in some capacity.  With these participants, we began by asking open-ended questions about their experience with saliency methods, such as ``\emph{What tasks do you use saliency methods for?}'', ``\emph{How do you decide which saliency method to use?}'', and ``\emph{What do you do if saliency methods disagree?}''. Next, we discussed each saliency method attribute.
We asked participants to describe if and how each attribute was important to their task and rank the attributes by importance. 
Finally, we walked users through example saliency cards (Fig.~\ref{fig:saliency-card}) and had users give us feedback on the design and usefulness.

The radiologist (\texttt{U9}) did not have experience using saliency methods but was interested in their application to medical decision-making.  
Since they were less familiar with ML, we structured this conversation slightly differently.  
We discussed five attributes\,---\,determinism, hyperparameter dependence, semantic directness, minimality, and perceptual correspondence. 
For each, we provided a definition, showed radiology examples demonstrating its implications (e.g., saliency with and without minimality), and discussed if and when it would be important to them in a clinical setting.

We conducted 30--60min interviews via video chat and compensated users with \$30 Amazon gift cards.
Our study received an IRB exemption from our organization. 
We obtained informed consent from participants, stored user data securely, and anonymized user details in the paper. 
See Sec. A.3 for additional study details.

\subsection{Saliency Cards Help Users Select Task-Appropriate Saliency Methods}

Despite experience with a broad range of saliency methods, including LIME~\citep{ribeiro2016should} (\texttt{U3}--\texttt{U7}), vanilla gradients~\citep{erhan2009visualizing, simonyan2013deep} (\texttt{U5}, \texttt{U6}, \texttt{U8}), integrated gradients~\citep{sundararajan2017axiomatic} (\texttt{U4}--\texttt{U6}), Grad-CAM~\citep{selvaraju2017grad} (\texttt{U1}, \texttt{U2}), SHAP~\citep{lundberg2017unified} (\texttt{U3}--\texttt{U5}), and SmoothGrad~\citep{smilkov2017smoothgrad} (\texttt{U5}), users chose saliency methods based on their popularity in prior work (\texttt{U1}--\texttt{U8}) and ease of implementation (\texttt{U2}, \texttt{U7}).
Users rarely considered algorithmic differences or evaluated the suitability of a particular method for their task. 
As a result, users were often unsure if their chosen saliency method was indeed appropriate for their task and worried that a different method could produce more accurate results.
Consequently, users wanted a more principled selection strategy based on formal evaluations but found extracting insight from existing documentation tedious.
As \texttt{U6} described, given that ``\textit{new methods come out every day}'' and ``\textit{reading all the papers is a difficult task that takes a lot of time,}'' even researchers find it challenging to acquire the knowledge needed to select saliency methods well-suited to their tasks. 

In contrast, we found that the attribute-based structure of saliency cards allowed users to more systematically select saliency methods based on properties important to their task.
Users prioritized each attribute based on their task requirements, experiences and preferences, and the expectations of their teammates.
For example, \texttt{U1} prioritized computational efficiency because their research requires them to compute saliency maps for every input in their dataset. 
An inefficient saliency method would be incompatible with their model evaluation process and prevent them from quickly iterating on model design choices.
\texttt{U9} prioritized determinism based on their personal experience. 
They were uncomfortable interpreting non-deterministic saliency maps because they do not encounter non-determinism in other medical technologies.
When prioritizing attributes, users in applied domains also needed to consider their teammates' expertise. 
For instance, \texttt{U7} prioritized semantic directness to help them communicate results to business clients without ML experience.
After prioritizing attributes, users utilized the visual saliency card documentation to juxtapose attribute summaries and pick a well-suited saliency method in just a few minutes.

Our user study also revealed that user priorities often differ or conflict\,---\,a surprising finding given that existing evaluations are often framed as ``tests'' every saliency method should pass~\cite{sundararajan2017axiomatic, adebayo2018sanity, ding2021evaluating}.
While one user would prioritize an attribute, another would deprioritize or explicitly not desire that attribute.
For example, \texttt{U8} prioritized minimality because they train machine learning models on long amino acid sequences, and their biochemist coworkers interpret the saliency method results. 
Without minimality, confirming the models have learned biologically meaningful features could require the biochemists to analyze the interactions between potentially hundreds of amino acids. 
However, \texttt{U4} explicitly preferred a less minimal method. 
They worried that a minimal saliency method might only highlight the features necessary for the model's prediction. 
Since they use saliency methods to manually analyze a few inputs, they want to view every feature relevant to the model's prediction to ensure their models do not learn spurious correlations.
Even users in similar roles had different priorities.
For instance, despite both being researchers who use saliency methods to analyze model behavior, \texttt{U2} and \texttt{U6} viewed the importance of input sensitivity differently.
\texttt{U2} regularly tests models by perturbing background features, so without input sensitivity, a saliency method could incorrectly assign importance to changes the model considers unimportant.
On the other hand, \texttt{U6} did not care about input sensitivity because they only use in-distribution data and do not worry about noise or perturbations impacting the inputs.
The frequency of conflicting priorities suggests there is not an ideal saliency method for every user and task. 
Thus, documentation is crucial to help users find a saliency method appropriate for their use case.

\subsection{Saliency Card Attributes Provide a Detailed Vocabulary for Discussing Saliency Methods}
Saliency cards provide a more precise attribute-based vocabulary that helps users communicate about saliency methods.
At the start of our interviews, participants often cited \textit{faithfulness} as an ideal attribute of saliency methods.
Faithfulness broadly refers to a saliency method's ability to reflect model reasoning accurately and correlates with the saliency card's sensitivity attributes.
However, after discussing the ten saliency card attributes, users had a more detailed language to describe saliency method characteristics.
For example, \texttt{U5} initially expected all saliency methods to achieve faithfulness. 
However, after working with saliency cards, \texttt{U5} more precisely articulated that they expected saliency methods to be label and model sensitive.
They did not care about a method's input sensitivity, even though it is typically considered part of faithfulness.
As a saliency method developer, \texttt{U5} needs to be able to communicate their exact design goals so users can understand the benefits, limitations, and appropriate use cases of the saliency method.
If they described their saliency method as faithful, users could incorrectly assume it is input sensitive, deploy it in an inappropriate setting, and misinterpret the results. 
Using a shared attribute-based vocabulary, users and developers can better communicate about a saliency method's specific attributes, evaluative results, and prescribed use cases.

The saliency card attributes also helped lay users discuss saliency methods. 
Before our user study, \texttt{U9} (a radiologist) had little experience with machine learning and was entirely unfamiliar with saliency methods.
However, by using the vocabulary of saliency card attributes, our conversation revealed differences in their expressed needs and expectations in the literature about what lay users want in a saliency method.
For example, minimality is often considered an essential attribute because it makes the visual saliency map easier to interpret~\citep{smilkov2017smoothgrad, kapishnikov2019xrai, sundararajan2019exploring}.
However, \texttt{U9} did not expect a saliency method to be minimal because they were accustomed to using noise in medical imaging to attenuate measurement uncertainty.
Using the saliency card attributes gave \texttt{U9} terminology they could use to communicate with ML experts and software vendors in charge of developing and deploying saliency methods.
Without this language with which to communicate, radiologists might not as deeply engage in the deployment process, leaving ML experts to rely on incorrect assumptions about radiologists' expectations.
However, with direct channels of communication, ML experts could work with radiologists to increase transparency in the deployment process, ensure they interpret saliency method results appropriately, and, even, develop new saliency methods explicitly designed for clinical imaging settings.

\subsection{Saliency Cards Inspire Areas for Future Work and New Documentation Practices}
\label{sec:user-study-future-work}

The attribute summaries led users to ask new questions about evaluating saliency methods and to hypothesize future research directions. 
By documenting evaluation results for a saliency method, saliency cards reveal that particular attributes and methods have been more heavily evaluated than others. 
For instance, comparing the saliency cards for integrated gradients~\citep{sundararajan2017axiomatic} (Fig. A2) and Grad-CAM~\citep{selvaraju2017grad} (Fig.~\ref{fig:saliency-card}) reveals that integrated gradients has been more rigorously tested for input sensitivity. 
Whereas previously, users would have had to extract evaluative results from multiple academic papers, saliency cards surface these discrepancies directly, inspiring users to hypothesize about Grad-CAM's performance on missing evaluations and express interest in completing the testing suite. 
Further, by categorizing individual evaluations, saliency cards expose that evaluations for the same attribute have varying testing strategies, such as testing meaningful~\citep{alvares2018towards} vs. noisy perturbations~\citep{kindermans2019reliability} or focusing on images~\citep{gomez2022metrics} vs. natural language modalities~\citep{ding2021evaluating}. 
Users were surprised to see the evaluation diversity, leading them to hypothesize new evaluation measures. 
For instance, some users were intrigued to run perceptibility tests on their data and models. 
As \texttt{U5} put it, ``\textit{If I have a specific use case in mind, I want to see the metrics on that specific use case.}'' 
They brainstormed ideas about integrating saliency cards into a suite of evaluations that generate customized saliency cards based on the user's model and datasets.

Inspecting some attributes revealed limitations of saliency cards and existing evaluations. 
Saliency cards group evaluations into user-centric attributes, but some attributes are challenging to test accurately. 
During our user study, \texttt{U8} was skeptical that existing evaluations appropriately assessed model sensitivity. 
Model sensitivity evaluations test that a saliency method responds to meaningful model changes, but \texttt{U8} argued that it is almost impossible to guarantee that a change to a black-box model is meaningful. 
For instance, a standard model sensitivity test measures the saliency method's response to layer randomization, but layer randomization might not be meaningful if that layer is redundant. 
In that case, layer randomization tests could incorrectly punish a model for not responding to an insignificant change. 
This issue might be solved as additional research invents new evaluations, including model sensitivity tests. 
However, it could also be that some attributes, like model sensitivity, are too broad. 
Perhaps breaking model sensitivity down into more precise categorizations, like layer randomization sensitivity, would provide more straightforward documentation.
Similarly, we expect the methodology attributes to evolve from open-ended descriptions to more consistent reports. 
For instance, the vocabulary used to describe computational efficiency may vary across saliency developers and research areas based on typical computing resources and dataset sizes.
As more saliency methods are documented and more evaluations are developed, we expect the saliency card attributes and their descriptions will evolve to better characterize saliency methods, facilitate cross-card comparison, and communicate with users.

Saliency method developers were inspired to document their methods with saliency cards and hoped consistent and thorough documentation would increase method adoption. 
Good documentation can make saliency methods easier to understand and use, ``\textit{If you want people to use your method, your need to have them understand it.}'' [\texttt{U8}]. 
Currently, saliency method developers have to generate documentation content that ranges from novel algorithmic decisions and implications in the paper to implementation considerations in the public code repository. 
This process can feel unprincipled, so developers were excited to have a template that fully captured critical considerations. 
For example, when developing their saliency method, \texttt{U8} documented their method's computational efficiency and hyperparameter dependence in their code repository, explaining ``\textit{We tried to make our documentation accessible to users. I tried to do some of this, but in an ad hoc way, and I didn't hit all of these [attributes].}'' 
They looked forward to adding additional documentation and making a saliency card for their method.

\section{Discussion and Limitations}
We present saliency cards, transparency documentation to describe, communicate, and compare saliency methods.
While documentation in other parts of the machine learning pipeline has led to increased trust and appropriate use~\citep{mcmillan2021reusable, huggingfacemodelcard, gebru2018datasheets, mitchell2019model}, saliency methods do not have documentation standards. 
As a result, users we interviewed struggled to stay informed with the ever-increasing number of saliency methods, forcing them to choose saliency methods based on popularity instead of a thorough understanding of their benefits and limitations. 
In response, saliency cards characterize saliency methods based on ten user-centric attributes that describe important usage considerations.
The saliency card attributes span different phases of the interpretation workflow, such as the saliency method's algorithmic properties, relationship to the model and data, and perceptibility by an end-user.
We evaluate saliency cards in a user study with nine participants, ranging from radiologists with limited knowledge of machine learning to saliency method developers.
With saliency cards, users prioritized attributes based on their task requirements, personal experience, and the expectations of their teammates, allowing them to select a saliency method appropriate for their needs and properly interpret its results. 
Further, the saliency card attributes provided users with a shared vocabulary to describe their needs and communicate about saliency methods without requiring extensive machine learning expertise.

Building saliency cards allowed us to analyze the research landscape, revealing areas for future work, such as task-specific saliency methods and evaluation metrics for under-evaluated attributes.
By documenting and comparing the methodological attributes of various saliency methods (Table A1), we identify the potential for new saliency methods that meet specific user priorities and future studies on the latent relationships between attributes.
Current saliency methods cannot achieve specific combinations of attributes.
For example, none of the saliency methods we surveyed were model agnostic and computationally efficient because model agnosticism is commonly achieved through expensive repeated perturbations.
Model agnosticism and computational efficiency were priorities for \texttt{U6} and \texttt{U7}, but currently, they must sacrifice one when choosing a saliency method.
New research could explore this gap, and others, by designing novel saliency methods that attain model agnosticism without forfeiting computational efficiency or proving that they are inexplicably inversely correlated. 

Saliency cards also revealed gaps in evaluation research, including under-evaluated attributes and saliency methods (Table A2).
For example, by compiling evaluative metrics for each attribute, we uncovered that there is far less research into how to measure perceptual correspondence, relative to other attributes such as input or model sensitivity.  
By identifying this gap, saliency cards prompt further research into how we might measure perceptual correspondence.  Better understanding how people perceive saliency maps could then motivate the design of new saliency visualizations\,---\,e.g., that expand static heatmaps by dynamically overlaying multiple attributions~\citep{olah2018building} to explicitly communicate limitations and preemptively avoid implying unwarranted signal. 
Table A2 also reveals that some saliency methods (e.g., SIS~\citep{carter2019made}) have been evaluated less than others (e.g., integrated gradients~\citep{sundararajan2017axiomatic}).
While the sensitivity and perceptibility attributes report results from existing evaluations, evaluation papers typically only test a subset of existing saliency methods. 
As a result, our users found it challenging to compare saliency methods evaluated on different tests. 
Future work could run missing evaluations or design test suites that report a saliency method's results on existing tests.

We intend saliency cards to be \textit{living artifacts} that start a conversation around saliency method documentation. 
To facilitate living documentation, we provide a public repository\footnote{\url{https://github.com/mitvis/saliency-cards}} containing saliency card templates, summaries of evaluations, and saliency cards for existing methods.
The repository serves as a centralized location for users to reference saliency methods.
As new saliency methods are developed to fulfill specific user needs, new saliency cards can be added to the repository.
Existing saliency cards can be continually updated with additional evaluative results stemming from new evaluation metrics and the application of existing metrics to unevaluated saliency methods.
As the saliency card repertoire expands, saliency card documentation will simultaneously evolve to support additional user needs.
As signaled in our user studies, new evaluations may reveal that some saliency card attributes are too broad and need to be decomposed into constituent attributes that more precisely articulate the evaluative takeaways. 
Likewise, new attributes or categories may emerge as more users from various backgrounds begin to use saliency methods and communicate their task-specific priorities.
By documenting saliency methods, we hope saliency cards support the continued rapid growth of saliency method research and evolve as needed alongside new developments.

\begin{acks}
This work is supported by NSF Award \#1900991, and via a grant from the MIT-IBM Watson AI Lab. H.S. was supported by the Kerr Fellowship.
Research was also sponsored by the United States Air Force Research Laboratory and the United States Air Force Artificial Intelligence Accelerator and was accomplished under Cooperative Agreement Number FA8750-19-2-1000. The views and conclusions contained in this document are those of the authors and should not be interpreted as representing the official policies, either expressed or implied, of the United States Air Force or the U.S. Government. The U.S. Government is authorized to reproduce and distribute reprints for Government purposes notwithstanding any copyright notation herein.
\end{acks}

\bibliographystyle{ACM-Reference-Format}
\bibliography{main}

\onecolumn
\appendix
\setcounter{table}{0}
\setcounter{figure}{0}
\renewcommand{\thetable}{A\arabic{table}}
\renewcommand{\thefigure}{A\arabic{figure}}

\clearpage

\section{Appendix}

\subsection{Additional Saliency Cards}
\label{sup:saliency-cards}
\begin{figure*}[ht]
  \centering
  \includegraphics[width=\textwidth]{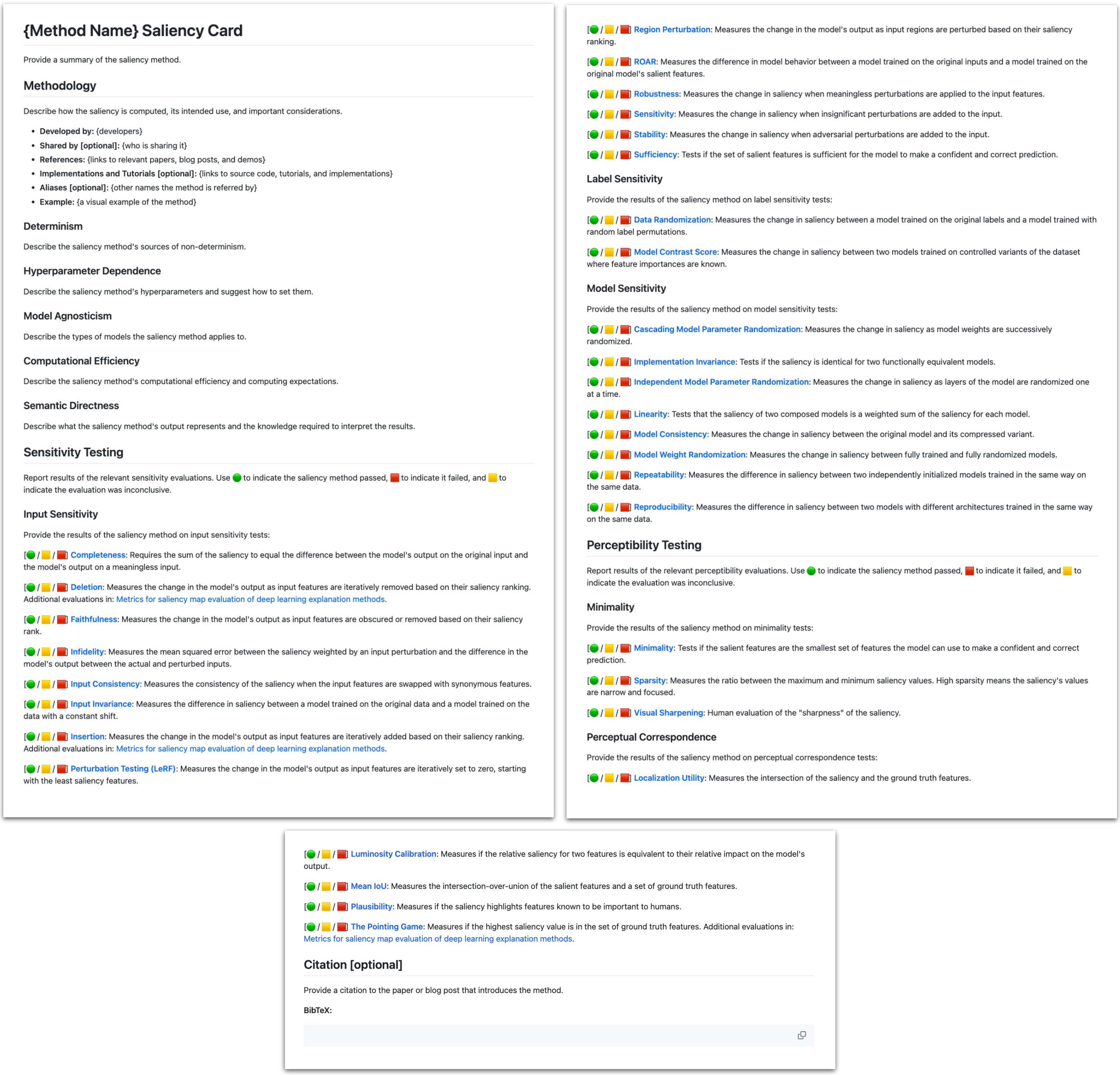}
  \caption{The saliency card template provided at \url{https://github.com/mitvis/saliency-cards}. The template describes what to include in each section of the saliency card and summarizes existing evaluations for sensitivity and perceptibility attributes.}
  \label{fig:saliency-card-template}
  \Description{Three pages of the saliency card template.}
\end{figure*}

\begin{figure*}[ht]
  \centering
  \includegraphics[width=\textwidth]{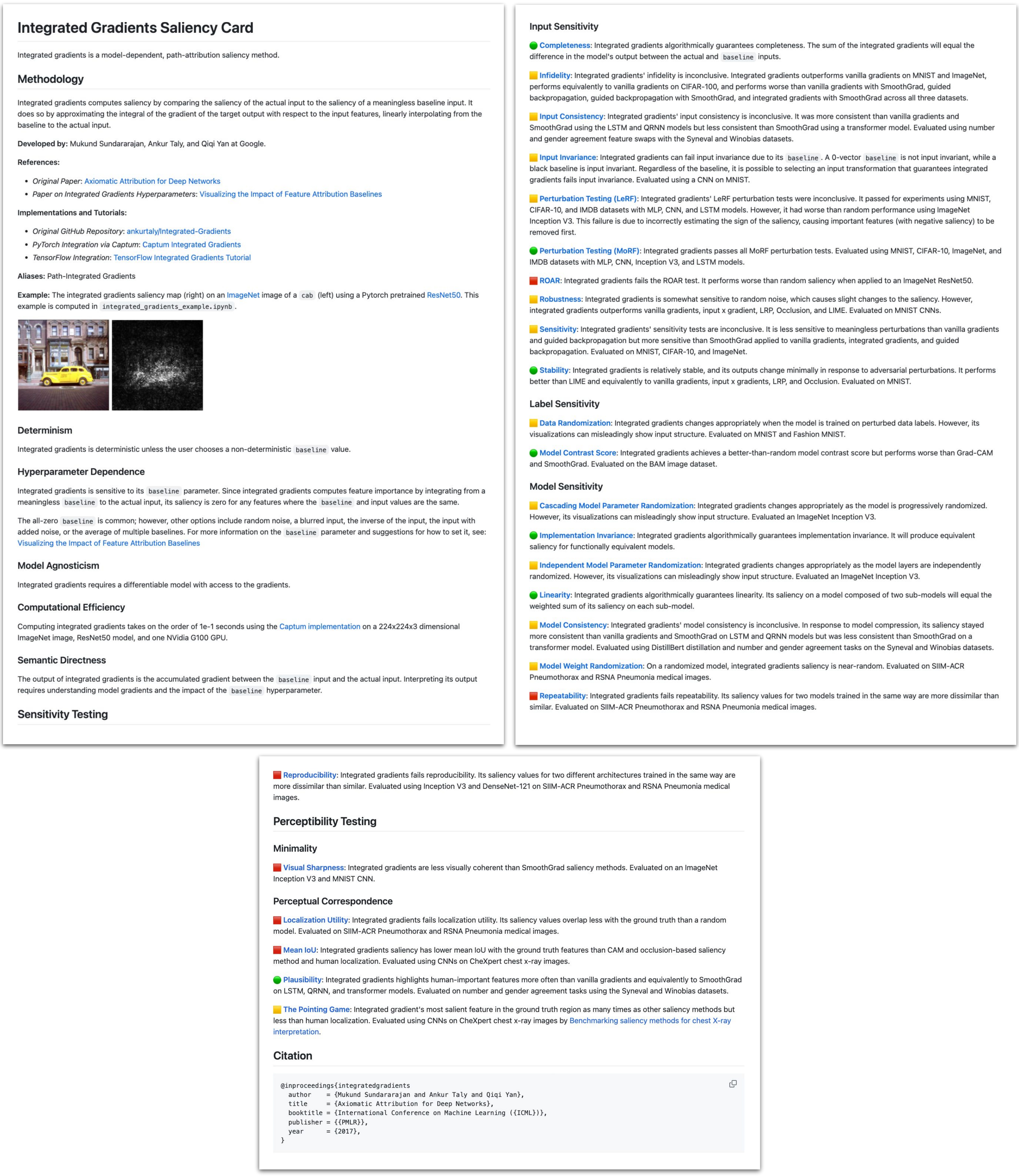}
  \caption{The saliency card for integrated gradients~\citep{sundararajan2017axiomatic} provided at \url{https://github.com/mitvis/saliency-cards}.}
  \label{fig:ig-saliency-card}
  \Description{Three images show pages 1--3 of the integrated gradients saliency card.}
\end{figure*}

The saliency cards repository (\url{https://github.com/mitvis/saliency-cards}) contains a saliency card template (Fig.~\ref{fig:saliency-card-template}) and example saliency cards, including Grad-CAM (Fig.~\ref{fig:saliency-card}) and integrated gradients (Fig.~\ref{fig:ig-saliency-card}). 
As more saliency methods are documented, developed, and evaluated, we expect the repository to serve as a centralized location for saliency documentation.

 \clearpage

\subsection{Saliency Card Attribute Comparison}
\label{sup:saliency-card-comparison}
Saliency cards expose differences in the methodology, sensitivity, and perceptibility of saliency methods.
We have provided example saliency cards for Grad-CAM~\citep{selvaraju2017grad} (Fig.~\ref{fig:saliency-card}) and integrated gradients~\citep{sundararajan2017axiomatic} (Fig.~\ref{fig:ig-saliency-card}) that describe their attributes.
Here, we compare the saliency card attributes for additional saliency methods, to analyze the research landscape and reveal gaps and opprotunities for future work.

\subsubsection{Methodology Comparison}
In Table~\ref{tab:methodology}, we compare the methodological attributes of 11 saliency methods.
We extract each method's determinism, hyperparameter dependence, model agnosticism, and semantic directness from its original paper and we compute each method's computational efficiency.

We compute computational efficiency of each method on a 224x224x3 ImageNet~\citep{deng2009imagenet} image and a pretrained PyTorch~\citep{paszke2019pytorch} ResNet50~\citep{he2016deep} using one Nvidia G100 GPU.
For consistency, we use each method's default parameters.
To account for noise in our timing procedure, we report the order of magnitude of the computation time as opposed to the raw value.
The results of the computational efficiency tests can be found at: \url{https://github.com/mitvis/saliency-cards/blob/main/paper_results/computational_efficiency.ipynb}.

For vanilla gradients~\citep{erhan2009visualizing, simonyan2013deep}, guided backprop~\citep{springenberg2014striving}, Grad-CAM~\citep{selvaraju2017grad}, integrated gradients~\citep{sundararajan2017axiomatic}, input x gradient, and SHAP~\citep{lundberg2017unified}, we use their Captum~\citep{kokhlikyan2020captum} implementations.
For SmoothGrad~\citep{smilkov2017smoothgrad} and XRAI~\citep{kapishnikov2019xrai}, we use the public implementation from Google PAIR\footnote{https://github.com/PAIR-code/saliency}.
For RISE~\footnote{https://github.com/eclique/RISE}~\citep{petsiuk2018rise}, SIS~\citep{carter2019made}~\footnote{https://github.com/gifford-lab/overinterpretation/tree/master; https://github.com/google-research/google-research/blob/master/sufficient\_input\_subsets/sis.py}, and  LIME~\footnote{https://github.com/marcotcr/lime}~\citep{ribeiro2016should}, we use the public implementations provided by their authors.
To provide a consistent saliency method interface across the different implementations, we provide a wrapper for each method and visualization code at:  \url{https://github.com/mitvis/saliency-cards}.

This is just one way to test the computational efficiency of each saliency method.
We expect the relative computational efficiencies may change depending on the implementation, model architecture, parameter settings, and data modality.
Similarly, there may be theoretical computational efficiency bounds that can be derived for each method.
As additional experiments reveal new computational efficiency results, saliency cards can evolve to contain a comprehensive overview of each method's computational efficiency.

\subsubsection{Sensitivity and Perceptibility Comparison}
Table~\ref{tab:sensitivty-perceptibility} summarizes the results of 33 saliency method evaluations that span the sensitivity and perceptibility attributes. 
We categorize each test's results for every saliency method it tested. 
Based on what each evaluation paper reports, we group the results into ``pass'', ``fail'', and ``inconclusive''.

While we try to directly report the paper's claims about each saliency method, in some cases, these categorizations can be subjective. 
\begin{table*}[bht]
\caption{We compare the methodological attributes of 11 saliency methods to systematically analyze the research landscape to understand gaps and opportunities for future work.}
\label{tab:methodology}
\tiny
{\RaggedRight
\begin{tabular}{
>{}m{0.12\linewidth}m{0.12\linewidth}m{0.19\linewidth}m{0.14\linewidth}m{0.14\linewidth}m{0.17\linewidth}}
\toprule
 & 
{\scriptsize\textcolor[HTML]{d04387}{\textbf{Determinism}}}                         & 
{\scriptsize\textcolor[HTML]{d04387}{\textbf{Hyperparameter Dependence}}}              & 
{\scriptsize\textcolor[HTML]{d04387}{\textbf{Model Agnosticism}}}   & 
{\scriptsize\textcolor[HTML]{d04387}{\textbf{Computational Efficiency}}}              & 
{\scriptsize\textcolor[HTML]{d04387}{\textbf{Semantic Directness}}}         \\ \midrule
\textbf{Vanilla Gradients \citep{erhan2009visualizing, simonyan2013deep}} & \cellcolor[HTML]{FFFFFF}Deterministic. & None. & Requires a differentiable model with access to gradients. & On the order of $1\mathrm{e}{-2}$ seconds. & The magnitude of the change in the model's output given a small change to an input feature.\\
\midrule
\textbf{SmoothGrad \citep{smilkov2017smoothgrad}} & \cellcolor[HTML]{FFFFFF}Non-deterministic noise perturbations.  & Gaussian noise parameters; the number of samples to average over. & Applicable to any saliency method. & Adds a $\sim$20x time increase.  & The average saliency across noisy versions of the input. \\
\midrule
\textbf{Guided BackProp \citep{springenberg2014striving}} & \cellcolor[HTML]{FFFFFF}Deterministic unless using a non-deterministic saliency method. & Saliency method (typically vanilla gradients). & Requires a differentiable model with access to gradients. & On the order of $1\mathrm{e}{-2}$ seconds using vanilla gradients. & The output of another gradient-based saliency method only considering paths through the model with positive gradients. \\
\midrule
\textbf{Grad-CAM \citep{selvaraju2017grad}} & Deterministic. & Interpolation method to upsample with; choice of convolutional layer (typically the last convolutional layer). & Requires a differentiable model, access to the gradients, and a convolutional layer. & On the order of $1\mathrm{e}{-2}$ seconds. & The positive attributions of the gradient-weighted feature maps from an internal convolutional layer.\\
\midrule
\textbf{Integrated Gradients \citep{sundararajan2017axiomatic}} & Deterministic unless using a non-deterministic baseline. & Baseline value; integral approximation parameters. & Requires a differentiable model with access to gradients. & On the order of $1\mathrm{e}{-1}$ seconds. & The accumulated gradient between the baseline input and the actual input. \\ 
\midrule
\textbf{Input X Gradient} & \cellcolor[HTML]{FFFFFF}Deterministic. & None. & Requires a differentiable model and access to gradients. & On the order of $1\mathrm{e}{-2}$ seconds. & The input feature value weighted by the gradient. \\
\midrule
{\textbf{XRAI \citep{kapishnikov2019xrai}}}  & \cellcolor[HTML]{FFFFFF} {Deterministic unless using a non-deterministic saliency method or segmentation method.} & {Segmentation method; saliency method (typically integrated gradients).} & {Requires input features that can be meaningfully clustered (e.g., image pixels).} & On the order of $1\mathrm{e}{1}$ seconds. & {The input regions with the largest sum of feature attribution.} \\ 
\midrule
{\textbf{RISE \citep{petsiuk2018rise}}}  & \cellcolor[HTML]{FFFFFF} {Non-deterministic mask generation.} & {Masking value; mask generation parameters.} & {No requirements on the model or access to internals.} & {On the order of $1\mathrm{e}{-1}$ seconds.} & {The sum of input masks weighed by the model's confidence on the masked input.}\\ 
\midrule
\textbf{SIS \citep{carter2019made}}  & \cellcolor[HTML]{FFFFFF} Deterministically produces a set of explanations per input.  & Feature replacement values; model confidence threshold. & No requirements on the model or access to internals. & SIS: prohibitively slow. Batched Gradient SIS: On the order of $1\mathrm{e}{1}$ seconds. & The minimum set of pixels necessary for the model to confidently produce the same output. \\ 
\midrule
\textbf{LIME \citep{ribeiro2016should}} & \cellcolor[HTML]{FFFFFF}Non-deterministic perturbations. & Linear surrogate model and parameters; input perturbation parameters. & No requirements on the model or access to internals.  & {On the order of $1\mathrm{e}{1}$ seconds.} & The positively contributing features learned by a surrogate model trained to mimic the original model's local decision boundary for the input.\\
\midrule
\textbf{SHAP \citep{lundberg2017unified}}  & Non-deterministic coalition sampling.  & Feature replacement values; linear model parameterization; regularization parameter.  & No requirements on the model or access to internals.  & GradientSHAP: on the order of $1\mathrm{e}{-1}$ seconds. KernelSHAP: on the order of $1\mathrm{e}{1}$ seconds. & The impact of each input feature on the output as defined by Shapley values. \\
\bottomrule   
\end{tabular}
}
\end{table*}
For instance, if an evaluation compares two saliency methods and one outperforms the other, it can be hard to evaluate which methods pass or fail the test. 
It could be just the top-performing one passes, both pass, or both fail. 
In evaluations that provide negative and positive controls (e.g., random baselines), we categorize each method based on those controls.

These categorizations are helpful for analyzing the gaps in the design space (e.g., limited evaluations for particular models or attributes); however, saliency cards must also provide a textual summary of the evaluation's results. 
This description helps mitigate the subjectivity of pass/fail/inconclusive categorizations and can provide helpful context to a user. 
For instance, seeing a lack of experimental controls could inspire a user to run additional baseline evaluations for an existing test or design new metrics that better separate the behavior of the saliency methods.

\begin{table*}[htp]
\centering
\caption{We summarize the results of 33 saliency method evaluations that describe the saliency card sensitivity and perceptibility attributes. This summarization reveals the need for additional evaluation using existing metrics and new metrics to further test under-evaluated attributes. We generalize the results of each test based on if they passed the evaluation (green \cmark), failed the evaluation (red \xmark), performed inconclusively (yellow ---), or were not tested (grey cell).}
\label{tab:sensitivty-perceptibility}
\tiny
\resizebox{\linewidth}{!}{
\begin{tabular}{llccccccccccccccccccccccccc}\toprule
\textbf{} &\textbf{} &\textbf{\rot{Vanilla Gradients (VG) \citep{erhan2009visualizing, simonyan2013deep}}} &\textbf{\rot{VG + SmoothGrad \citep{smilkov2017smoothgrad}}} &\textbf{\rot{Guided BackProp (GBP) \citep{springenberg2014striving}}} &\textbf{\rot{Grad-CAM \citep{selvaraju2017grad}}} &\textbf{\rot{Guided Grad-CAM \citep{selvaraju2017grad}}} &\textbf{\rot{Integrated Gradients (IG) \citep{sundararajan2017axiomatic}}} &\textbf{\rot{IG + SmoothGrad \citep{sundararajan2017axiomatic, smilkov2017smoothgrad}}} &\textbf{\rot{Gradient $\cdot$ Input}} &\textbf{\rot{XRAI \citep{kapishnikov2019xrai}}} &\textbf{\rot{DeepLift \citep{shrikumar2017learning}}} &\textbf{\rot{LRP \citep{li2019beyond}}} &\textbf{\rot{Ablation CAM \citep{ramaswamy2020ablation}}} &\textbf{\rot{Score-CAM \citep{wang2020score}}} &\textbf{\rot{RISE \citep{petsiuk2018rise}}} &\textbf{\rot{Grad-CAM++ \citep{chattopadhay2018grad}}} &\textbf{\rot{SIS \citep{carter2019made}}} &\textbf{\rot{Deconvnet \citep{zeiler2014visualizing}}} &\textbf{\rot{CAM \citep{zhou2016learning}}} &\textbf{\rot{Occlusion \citep{zeiler2014visualizing}}} &\textbf{\rot{Eigen-CAM \citep{muhammad2020eigen}}} &\textbf{\rot{LIME \citep{ribeiro2016should}}} &\textbf{\rot{GBP + SmoothGrad \citep{springenberg2014striving, smilkov2017smoothgrad}}} &\textbf{\rot{SHAP \citep{lundberg2017unified}}} &\textbf{\rot{Deep Taylor Decomposition \citep{montavon2017explaining}}} &\textbf{\rot{PatternNet \citep{kindermans2017learning}}} \\\midrule
{\scriptsize\textcolor[HTML]{4aa5ea}{\textbf{Input Sensitivity}}}  &\textbf{Completeness \citep{sundararajan2017axiomatic} } &\cellcolor[HTML]{f3f3f3} &\cellcolor[HTML]{f3f3f3} &\cellcolor[HTML]{f3f3f3} &\cellcolor[HTML]{f3f3f3} &\cellcolor[HTML]{f3f3f3} &\cellcolor[HTML]{d9ead3}\cmark &\cellcolor[HTML]{f3f3f3} &\cellcolor[HTML]{f3f3f3} &\cellcolor[HTML]{f3f3f3} &\cellcolor[HTML]{d9ead3}\cmark &\cellcolor[HTML]{d9ead3}\cmark &\cellcolor[HTML]{f3f3f3} &\cellcolor[HTML]{f3f3f3} &\cellcolor[HTML]{f3f3f3} &\cellcolor[HTML]{f3f3f3} &\cellcolor[HTML]{f3f3f3} &\cellcolor[HTML]{f3f3f3} &\cellcolor[HTML]{f3f3f3} &\cellcolor[HTML]{f3f3f3} &\cellcolor[HTML]{f3f3f3} &\cellcolor[HTML]{f3f3f3} &\cellcolor[HTML]{f3f3f3} &\cellcolor[HTML]{f3f3f3} &\cellcolor[HTML]{f3f3f3} &\cellcolor[HTML]{f3f3f3} \\
&\textbf{Deletion \citep{petsiuk2018rise}} &\cellcolor[HTML]{f3f3f3} &\cellcolor[HTML]{f3f3f3} &\cellcolor[HTML]{f3f3f3} &\cellcolor[HTML]{ffecb3}--- &\cellcolor[HTML]{f3f3f3} &\cellcolor[HTML]{f3f3f3} &\cellcolor[HTML]{f3f3f3} &\cellcolor[HTML]{f3f3f3} &\cellcolor[HTML]{f3f3f3} &\cellcolor[HTML]{f3f3f3} &\cellcolor[HTML]{f3f3f3} &\cellcolor[HTML]{f3f3f3} &\cellcolor[HTML]{f3f3f3} &\cellcolor[HTML]{d9ead3}\cmark &\cellcolor[HTML]{f3f3f3} &\cellcolor[HTML]{f3f3f3} &\cellcolor[HTML]{f3f3f3} &\cellcolor[HTML]{f3f3f3} &\cellcolor[HTML]{f3f3f3} &\cellcolor[HTML]{f3f3f3} &\cellcolor[HTML]{ffecb3}--- &\cellcolor[HTML]{f3f3f3} &\cellcolor[HTML]{f3f3f3} &\cellcolor[HTML]{f3f3f3} &\cellcolor[HTML]{f3f3f3} \\
&\textbf{Faithfulness \citep{alvares2018towards}} &\cellcolor[HTML]{f3f3f3} &\cellcolor[HTML]{f3f3f3} &\cellcolor[HTML]{f3f3f3} &\cellcolor[HTML]{f3f3f3} &\cellcolor[HTML]{f3f3f3} &\cellcolor[HTML]{f3f3f3} &\cellcolor[HTML]{f3f3f3} &\cellcolor[HTML]{f3f3f3} &\cellcolor[HTML]{f3f3f3} &\cellcolor[HTML]{f3f3f3} &\cellcolor[HTML]{f3f3f3} &\cellcolor[HTML]{f3f3f3} &\cellcolor[HTML]{f3f3f3} &\cellcolor[HTML]{f3f3f3} &\cellcolor[HTML]{f3f3f3} &\cellcolor[HTML]{f3f3f3} &\cellcolor[HTML]{f3f3f3} &\cellcolor[HTML]{f3f3f3} &\cellcolor[HTML]{f3f3f3} &\cellcolor[HTML]{f3f3f3} &\cellcolor[HTML]{d9ead3}\cmark &\cellcolor[HTML]{f3f3f3} &\cellcolor[HTML]{d9ead3}\cmark &\cellcolor[HTML]{f3f3f3} &\cellcolor[HTML]{f3f3f3} \\
&\textbf{Infidelity \citep{yeh2019on}} &\cellcolor[HTML]{ffecb3}--- &\cellcolor[HTML]{d9ead3}\cmark &\cellcolor[HTML]{ffecb3}--- &\cellcolor[HTML]{f3f3f3} &\cellcolor[HTML]{f3f3f3} &\cellcolor[HTML]{ffecb3}--- &\cellcolor[HTML]{d9ead3}\cmark &\cellcolor[HTML]{f3f3f3} &\cellcolor[HTML]{f3f3f3} &\cellcolor[HTML]{f3f3f3} &\cellcolor[HTML]{f3f3f3} &\cellcolor[HTML]{f3f3f3} &\cellcolor[HTML]{f3f3f3} &\cellcolor[HTML]{f3f3f3} &\cellcolor[HTML]{f3f3f3} &\cellcolor[HTML]{f3f3f3} &\cellcolor[HTML]{f3f3f3} &\cellcolor[HTML]{f3f3f3} &\cellcolor[HTML]{f3f3f3} &\cellcolor[HTML]{f3f3f3} &\cellcolor[HTML]{f3f3f3} &\cellcolor[HTML]{d9ead3}\cmark &\cellcolor[HTML]{d9ead3}\cmark &\cellcolor[HTML]{f3f3f3} &\cellcolor[HTML]{f3f3f3} 
\\
&\textbf{Input Consistency \citep{ding2021evaluating}} &\cellcolor[HTML]{ffecb3}--- &\cellcolor[HTML]{ffecb3}--- &\cellcolor[HTML]{f3f3f3} &\cellcolor[HTML]{f3f3f3} &\cellcolor[HTML]{f3f3f3} &\cellcolor[HTML]{ffecb3}--- &\cellcolor[HTML]{f3f3f3} &\cellcolor[HTML]{f3f3f3} &\cellcolor[HTML]{f3f3f3} &\cellcolor[HTML]{f3f3f3} &\cellcolor[HTML]{f3f3f3} &\cellcolor[HTML]{f3f3f3} &\cellcolor[HTML]{f3f3f3} &\cellcolor[HTML]{f3f3f3} &\cellcolor[HTML]{f3f3f3} &\cellcolor[HTML]{f3f3f3} &\cellcolor[HTML]{f3f3f3} &\cellcolor[HTML]{f3f3f3} &\cellcolor[HTML]{f3f3f3} &\cellcolor[HTML]{f3f3f3} &\cellcolor[HTML]{f3f3f3} &\cellcolor[HTML]{f3f3f3} &\cellcolor[HTML]{f3f3f3} &\cellcolor[HTML]{f3f3f3} &\cellcolor[HTML]{f3f3f3} 
\\
&\textbf{Input Invariance \citep{kindermans2019reliability}} &\cellcolor[HTML]{d9ead3}\cmark &\cellcolor[HTML]{d9ead3}\cmark &\cellcolor[HTML]{d9ead3}\cmark &\cellcolor[HTML]{f3f3f3} &\cellcolor[HTML]{f3f3f3} &\cellcolor[HTML]{ffecb3}--- &\cellcolor[HTML]{d9ead3}\cmark &\cellcolor[HTML]{f4cccc}\xmark &\cellcolor[HTML]{f3f3f3} &\cellcolor[HTML]{f3f3f3} &\cellcolor[HTML]{f4cccc}\xmark &\cellcolor[HTML]{f3f3f3} &\cellcolor[HTML]{f3f3f3} &\cellcolor[HTML]{f3f3f3} &\cellcolor[HTML]{f3f3f3} &\cellcolor[HTML]{f3f3f3} &\cellcolor[HTML]{f3f3f3} &\cellcolor[HTML]{f3f3f3} &\cellcolor[HTML]{f3f3f3} &\cellcolor[HTML]{f3f3f3} &\cellcolor[HTML]{f3f3f3} &\cellcolor[HTML]{d9ead3}\cmark &\cellcolor[HTML]{f3f3f3} &\cellcolor[HTML]{ffecb3}--- &\cellcolor[HTML]{d9ead3}\cmark 
\\
&\textbf{Insertion \citep{petsiuk2018rise}} &\cellcolor[HTML]{f3f3f3} &\cellcolor[HTML]{f3f3f3} &\cellcolor[HTML]{f3f3f3} &\cellcolor[HTML]{ffecb3}--- &\cellcolor[HTML]{f3f3f3} &\cellcolor[HTML]{f3f3f3} &\cellcolor[HTML]{f3f3f3} &\cellcolor[HTML]{f3f3f3} &\cellcolor[HTML]{f3f3f3} &\cellcolor[HTML]{f3f3f3} &\cellcolor[HTML]{f3f3f3} &\cellcolor[HTML]{f3f3f3} &\cellcolor[HTML]{f3f3f3} &\cellcolor[HTML]{d9ead3}\cmark &\cellcolor[HTML]{f3f3f3} &\cellcolor[HTML]{f3f3f3} &\cellcolor[HTML]{f3f3f3} &\cellcolor[HTML]{f3f3f3} &\cellcolor[HTML]{f3f3f3} &\cellcolor[HTML]{f3f3f3} &\cellcolor[HTML]{ffecb3}--- &\cellcolor[HTML]{f3f3f3} &\cellcolor[HTML]{f3f3f3} &\cellcolor[HTML]{f3f3f3} &\cellcolor[HTML]{f3f3f3} 
\\
&\textbf{Perturbation Testing (LeRF) \citep{ancona2017unified}} &\cellcolor[HTML]{ffecb3}--- &\cellcolor[HTML]{f3f3f3} &\cellcolor[HTML]{f3f3f3} &\cellcolor[HTML]{f3f3f3} &\cellcolor[HTML]{f3f3f3} &\cellcolor[HTML]{ffecb3}--- &\cellcolor[HTML]{f3f3f3} &\cellcolor[HTML]{ffecb3}--- &\cellcolor[HTML]{f3f3f3} &\cellcolor[HTML]{ffecb3}--- &\cellcolor[HTML]{ffecb3}--- &\cellcolor[HTML]{f3f3f3} &\cellcolor[HTML]{f3f3f3} &\cellcolor[HTML]{f3f3f3} &\cellcolor[HTML]{f3f3f3} &\cellcolor[HTML]{f3f3f3} &\cellcolor[HTML]{f3f3f3} &\cellcolor[HTML]{f3f3f3} &\cellcolor[HTML]{ffecb3}--- &\cellcolor[HTML]{f3f3f3} &\cellcolor[HTML]{f3f3f3} &\cellcolor[HTML]{f3f3f3} &\cellcolor[HTML]{f3f3f3} &\cellcolor[HTML]{f3f3f3} &\cellcolor[HTML]{f3f3f3} 
\\
&\textbf{Perturbation Testing (MoRF) \citep{ancona2017unified}} &\cellcolor[HTML]{ffecb3}--- &\cellcolor[HTML]{f3f3f3} &\cellcolor[HTML]{f3f3f3} &\cellcolor[HTML]{f3f3f3} &\cellcolor[HTML]{f3f3f3} &\cellcolor[HTML]{d9ead3}\cmark &\cellcolor[HTML]{f3f3f3} &\cellcolor[HTML]{d9ead3}\cmark &\cellcolor[HTML]{f3f3f3} &\cellcolor[HTML]{d9ead3}\cmark &\cellcolor[HTML]{ffecb3}--- &\cellcolor[HTML]{f3f3f3} &\cellcolor[HTML]{f3f3f3} &\cellcolor[HTML]{f3f3f3} &\cellcolor[HTML]{f3f3f3} &\cellcolor[HTML]{f3f3f3} &\cellcolor[HTML]{f3f3f3} &\cellcolor[HTML]{f3f3f3} &\cellcolor[HTML]{d9ead3}\cmark &\cellcolor[HTML]{f3f3f3} &\cellcolor[HTML]{f3f3f3} &\cellcolor[HTML]{f3f3f3} &\cellcolor[HTML]{f3f3f3} &\cellcolor[HTML]{f3f3f3} &\cellcolor[HTML]{f3f3f3} 
\\
&\textbf{Region Perturbation \citep{samek2016evaluating}} &\cellcolor[HTML]{f3f3f3} &\cellcolor[HTML]{f3f3f3} &\cellcolor[HTML]{f3f3f3} &\cellcolor[HTML]{f3f3f3} &\cellcolor[HTML]{f3f3f3} &\cellcolor[HTML]{f3f3f3} &\cellcolor[HTML]{f3f3f3} &\cellcolor[HTML]{f3f3f3} &\cellcolor[HTML]{f3f3f3} &\cellcolor[HTML]{f3f3f3} &\cellcolor[HTML]{d9ead3}\cmark &\cellcolor[HTML]{f3f3f3} &\cellcolor[HTML]{f3f3f3} &\cellcolor[HTML]{f3f3f3} &\cellcolor[HTML]{f3f3f3} &\cellcolor[HTML]{f3f3f3} &\cellcolor[HTML]{d9ead3}\cmark &\cellcolor[HTML]{f3f3f3} &\cellcolor[HTML]{f3f3f3} &\cellcolor[HTML]{f3f3f3} &\cellcolor[HTML]{f3f3f3} &\cellcolor[HTML]{f3f3f3} &\cellcolor[HTML]{f3f3f3} &\cellcolor[HTML]{f3f3f3} &\cellcolor[HTML]{f3f3f3} \\
&\textbf{ROAR \citep{hooker2019benchmark}} &\cellcolor[HTML]{f4cccc}\xmark &\cellcolor[HTML]{ffecb3}--- &\cellcolor[HTML]{f4cccc}\xmark &\cellcolor[HTML]{f3f3f3} &\cellcolor[HTML]{f3f3f3} &\cellcolor[HTML]{f4cccc}\xmark &\cellcolor[HTML]{f3f3f3} &\cellcolor[HTML]{f3f3f3} &\cellcolor[HTML]{f3f3f3} &\cellcolor[HTML]{f3f3f3} &\cellcolor[HTML]{f3f3f3} &\cellcolor[HTML]{f3f3f3} &\cellcolor[HTML]{f3f3f3} &\cellcolor[HTML]{f3f3f3} &\cellcolor[HTML]{f3f3f3} &\cellcolor[HTML]{f3f3f3} &\cellcolor[HTML]{f3f3f3} &\cellcolor[HTML]{f3f3f3} &\cellcolor[HTML]{f3f3f3} &\cellcolor[HTML]{f3f3f3} &\cellcolor[HTML]{f3f3f3} &\cellcolor[HTML]{f3f3f3} &\cellcolor[HTML]{f3f3f3} &\cellcolor[HTML]{f3f3f3} &\cellcolor[HTML]{f3f3f3} 
\\
&\textbf{Robustness \citep{alvarez2018on}} &\cellcolor[HTML]{ffecb3}--- &\cellcolor[HTML]{f3f3f3} &\cellcolor[HTML]{f3f3f3} &\cellcolor[HTML]{f3f3f3} &\cellcolor[HTML]{f3f3f3} &\cellcolor[HTML]{ffecb3}--- &\cellcolor[HTML]{f3f3f3} &\cellcolor[HTML]{ffecb3}--- &\cellcolor[HTML]{f3f3f3} &\cellcolor[HTML]{f3f3f3} &\cellcolor[HTML]{ffecb3}--- &\cellcolor[HTML]{f3f3f3} &\cellcolor[HTML]{f3f3f3} &\cellcolor[HTML]{f3f3f3} &\cellcolor[HTML]{f3f3f3} &\cellcolor[HTML]{f3f3f3} &\cellcolor[HTML]{f3f3f3} &\cellcolor[HTML]{f3f3f3} &\cellcolor[HTML]{f4cccc}\xmark &\cellcolor[HTML]{f3f3f3} &\cellcolor[HTML]{f4cccc}\xmark &\cellcolor[HTML]{f3f3f3} &\cellcolor[HTML]{f4cccc}\xmark &\cellcolor[HTML]{f3f3f3} &\cellcolor[HTML]{f3f3f3} 
\\
&\textbf{Sensitivity \citep{yeh2019on}} &\cellcolor[HTML]{ffecb3}--- &\cellcolor[HTML]{d9ead3}\cmark &\cellcolor[HTML]{ffecb3}--- &\cellcolor[HTML]{f3f3f3} &\cellcolor[HTML]{f3f3f3} &\cellcolor[HTML]{ffecb3}--- &\cellcolor[HTML]{d9ead3}\cmark &\cellcolor[HTML]{f3f3f3} &\cellcolor[HTML]{f3f3f3} &\cellcolor[HTML]{f3f3f3} &\cellcolor[HTML]{f3f3f3} &\cellcolor[HTML]{f3f3f3} &\cellcolor[HTML]{f3f3f3} &\cellcolor[HTML]{f3f3f3} &\cellcolor[HTML]{f3f3f3} &\cellcolor[HTML]{f3f3f3} &\cellcolor[HTML]{f3f3f3} &\cellcolor[HTML]{f3f3f3} &\cellcolor[HTML]{f3f3f3} &\cellcolor[HTML]{f3f3f3} &\cellcolor[HTML]{f3f3f3} &\cellcolor[HTML]{d9ead3}\cmark &\cellcolor[HTML]{d9ead3}\cmark &\cellcolor[HTML]{f3f3f3} &\cellcolor[HTML]{f3f3f3} 
\\
&\textbf{Stability \citep{alvares2018towards}} &\cellcolor[HTML]{d9ead3}\cmark &\cellcolor[HTML]{f3f3f3} &\cellcolor[HTML]{f3f3f3} &\cellcolor[HTML]{f3f3f3} &\cellcolor[HTML]{f3f3f3} &\cellcolor[HTML]{d9ead3}\cmark &\cellcolor[HTML]{f3f3f3} &\cellcolor[HTML]{d9ead3}\cmark &\cellcolor[HTML]{f3f3f3} &\cellcolor[HTML]{f3f3f3} &\cellcolor[HTML]{d9ead3}\cmark &\cellcolor[HTML]{f3f3f3} &\cellcolor[HTML]{f3f3f3} &\cellcolor[HTML]{f3f3f3} &\cellcolor[HTML]{f3f3f3} &\cellcolor[HTML]{f3f3f3} &\cellcolor[HTML]{f3f3f3} &\cellcolor[HTML]{f3f3f3} &\cellcolor[HTML]{d9ead3}\cmark &\cellcolor[HTML]{f3f3f3} &\cellcolor[HTML]{f4cccc}\xmark &\cellcolor[HTML]{f3f3f3} &\cellcolor[HTML]{f4cccc}\xmark &\cellcolor[HTML]{f3f3f3} &\cellcolor[HTML]{f3f3f3}
\\
&\textbf{Sufficiency \citep{carter2019made}} &\cellcolor[HTML]{f3f3f3} &\cellcolor[HTML]{f3f3f3} &\cellcolor[HTML]{f3f3f3} &\cellcolor[HTML]{f3f3f3} &\cellcolor[HTML]{f3f3f3} &\cellcolor[HTML]{f3f3f3} &\cellcolor[HTML]{f3f3f3} &\cellcolor[HTML]{f3f3f3} &\cellcolor[HTML]{f3f3f3} &\cellcolor[HTML]{f3f3f3} &\cellcolor[HTML]{f3f3f3} &\cellcolor[HTML]{f3f3f3} &\cellcolor[HTML]{f3f3f3} &\cellcolor[HTML]{f3f3f3} &\cellcolor[HTML]{f3f3f3} &\cellcolor[HTML]{d9ead3}\cmark &\cellcolor[HTML]{f3f3f3} &\cellcolor[HTML]{f3f3f3} &\cellcolor[HTML]{f3f3f3} &\cellcolor[HTML]{f3f3f3} &\cellcolor[HTML]{f3f3f3} &\cellcolor[HTML]{f3f3f3} &\cellcolor[HTML]{f3f3f3} &\cellcolor[HTML]{f3f3f3} &\cellcolor[HTML]{f3f3f3} \\
\midrule
{\scriptsize\textcolor[HTML]{4aa5ea}{\textbf{Label Sensitivity}}} &\textbf{Data Randomization \citep{adebayo2018sanity}} &\cellcolor[HTML]{d9ead3}\cmark &\cellcolor[HTML]{d9ead3}\cmark &\cellcolor[HTML]{ffecb3}--- &\cellcolor[HTML]{d9ead3}\cmark &\cellcolor[HTML]{ffecb3}--- &\cellcolor[HTML]{ffecb3}--- &\cellcolor[HTML]{ffecb3}--- &\cellcolor[HTML]{ffecb3}--- &\cellcolor[HTML]{f3f3f3} &\cellcolor[HTML]{f3f3f3} &\cellcolor[HTML]{f3f3f3} &\cellcolor[HTML]{f3f3f3} &\cellcolor[HTML]{f3f3f3} &\cellcolor[HTML]{f3f3f3} &\cellcolor[HTML]{f3f3f3} &\cellcolor[HTML]{f3f3f3} &\cellcolor[HTML]{f3f3f3} &\cellcolor[HTML]{f3f3f3} &\cellcolor[HTML]{f3f3f3} &\cellcolor[HTML]{f3f3f3} &\cellcolor[HTML]{f3f3f3} &\cellcolor[HTML]{f3f3f3} &\cellcolor[HTML]{f3f3f3} &\cellcolor[HTML]{f3f3f3} &\cellcolor[HTML]{f3f3f3} \\
&\textbf{Model Contrast Score \citep{yang2019benchmarking}} &\cellcolor[HTML]{ffecb3}--- &\cellcolor[HTML]{ffecb3}--- &\cellcolor[HTML]{ffecb3}--- &\cellcolor[HTML]{d9ead3}\cmark &\cellcolor[HTML]{ffecb3}--- &\cellcolor[HTML]{d9ead3}\cmark &\cellcolor[HTML]{ffecb3}--- &\cellcolor[HTML]{ffecb3}--- &\cellcolor[HTML]{f3f3f3} &\cellcolor[HTML]{f3f3f3} &\cellcolor[HTML]{f3f3f3} &\cellcolor[HTML]{f3f3f3} &\cellcolor[HTML]{f3f3f3} &\cellcolor[HTML]{f3f3f3} &\cellcolor[HTML]{f3f3f3} &\cellcolor[HTML]{f3f3f3} &\cellcolor[HTML]{f3f3f3} &\cellcolor[HTML]{f3f3f3} &\cellcolor[HTML]{f3f3f3} &\cellcolor[HTML]{f3f3f3} &\cellcolor[HTML]{f3f3f3} &\cellcolor[HTML]{f3f3f3} &\cellcolor[HTML]{f3f3f3} &\cellcolor[HTML]{f3f3f3} &\cellcolor[HTML]{f3f3f3} \\
\midrule
{\scriptsize\textcolor[HTML]{4aa5ea}{\textbf{Model Sensitivity}}} &\textbf{Cascading Model Randomization \citep{adebayo2018sanity}} &\cellcolor[HTML]{d9ead3}\cmark &\cellcolor[HTML]{d9ead3}\cmark &\cellcolor[HTML]{f4cccc}\xmark &\cellcolor[HTML]{d9ead3}\cmark &\cellcolor[HTML]{f4cccc}\xmark &\cellcolor[HTML]{ffecb3}--- &\cellcolor[HTML]{ffecb3}--- &\cellcolor[HTML]{ffecb3}--- &\cellcolor[HTML]{f3f3f3} &\cellcolor[HTML]{f3f3f3} &\cellcolor[HTML]{f3f3f3} &\cellcolor[HTML]{f3f3f3} &\cellcolor[HTML]{f3f3f3} &\cellcolor[HTML]{f3f3f3} &\cellcolor[HTML]{f3f3f3} &\cellcolor[HTML]{f3f3f3} &\cellcolor[HTML]{f3f3f3} &\cellcolor[HTML]{f3f3f3} &\cellcolor[HTML]{f3f3f3} &\cellcolor[HTML]{f3f3f3} &\cellcolor[HTML]{f3f3f3} &\cellcolor[HTML]{f3f3f3} &\cellcolor[HTML]{f3f3f3} &\cellcolor[HTML]{f3f3f3} &\cellcolor[HTML]{f3f3f3} \\
&\textbf{Implementation Invariance \citep{sundararajan2017axiomatic}} &\cellcolor[HTML]{d9ead3}\cmark &\cellcolor[HTML]{f3f3f3} &\cellcolor[HTML]{f3f3f3} &\cellcolor[HTML]{f3f3f3} &\cellcolor[HTML]{f3f3f3} &\cellcolor[HTML]{d9ead3}\cmark &\cellcolor[HTML]{f3f3f3} &\cellcolor[HTML]{f3f3f3} &\cellcolor[HTML]{f3f3f3} &\cellcolor[HTML]{f4cccc}\xmark &\cellcolor[HTML]{f4cccc}\xmark &\cellcolor[HTML]{f3f3f3} &\cellcolor[HTML]{f3f3f3} &\cellcolor[HTML]{f3f3f3} &\cellcolor[HTML]{f3f3f3} &\cellcolor[HTML]{f3f3f3} &\cellcolor[HTML]{f3f3f3} &\cellcolor[HTML]{f3f3f3} &\cellcolor[HTML]{f3f3f3} &\cellcolor[HTML]{f3f3f3} &\cellcolor[HTML]{f3f3f3} &\cellcolor[HTML]{f3f3f3} &\cellcolor[HTML]{f3f3f3} &\cellcolor[HTML]{f3f3f3} &\cellcolor[HTML]{f3f3f3} \\
&\textbf{Independent Model Randomization \citep{adebayo2018sanity}} &\cellcolor[HTML]{d9ead3}\cmark &\cellcolor[HTML]{d9ead3}\cmark &\cellcolor[HTML]{f4cccc}\xmark &\cellcolor[HTML]{d9ead3}\cmark &\cellcolor[HTML]{f4cccc}\xmark &\cellcolor[HTML]{ffecb3}--- &\cellcolor[HTML]{ffecb3}--- &\cellcolor[HTML]{ffecb3}--- &\cellcolor[HTML]{f3f3f3} &\cellcolor[HTML]{f3f3f3} &\cellcolor[HTML]{f3f3f3} &\cellcolor[HTML]{f3f3f3} &\cellcolor[HTML]{f3f3f3} &\cellcolor[HTML]{f3f3f3} &\cellcolor[HTML]{f3f3f3} &\cellcolor[HTML]{f3f3f3} &\cellcolor[HTML]{f3f3f3} &\cellcolor[HTML]{f3f3f3} &\cellcolor[HTML]{f3f3f3} &\cellcolor[HTML]{f3f3f3} &\cellcolor[HTML]{f3f3f3} &\cellcolor[HTML]{f3f3f3} &\cellcolor[HTML]{f3f3f3} &\cellcolor[HTML]{f3f3f3} &\cellcolor[HTML]{f3f3f3} \\
&\textbf{Linearity \citep{sundararajan2017axiomatic}} &\cellcolor[HTML]{f3f3f3} &\cellcolor[HTML]{f3f3f3} &\cellcolor[HTML]{f3f3f3} &\cellcolor[HTML]{f3f3f3} &\cellcolor[HTML]{f3f3f3} &\cellcolor[HTML]{d9ead3}\cmark &\cellcolor[HTML]{f3f3f3} &\cellcolor[HTML]{f3f3f3} &\cellcolor[HTML]{f3f3f3} &\cellcolor[HTML]{f3f3f3} &\cellcolor[HTML]{f3f3f3} &\cellcolor[HTML]{f3f3f3} &\cellcolor[HTML]{f3f3f3} &\cellcolor[HTML]{f3f3f3} &\cellcolor[HTML]{f3f3f3} &\cellcolor[HTML]{f3f3f3} &\cellcolor[HTML]{f3f3f3} &\cellcolor[HTML]{f3f3f3} &\cellcolor[HTML]{f3f3f3} &\cellcolor[HTML]{f3f3f3} &\cellcolor[HTML]{f3f3f3} &\cellcolor[HTML]{f3f3f3} &\cellcolor[HTML]{f3f3f3} &\cellcolor[HTML]{f3f3f3} &\cellcolor[HTML]{f3f3f3}
\\
&\textbf{Model Consistency \citep{ding2021evaluating}} &\cellcolor[HTML]{ffecb3}--- &\cellcolor[HTML]{ffecb3}--- &\cellcolor[HTML]{f3f3f3} &\cellcolor[HTML]{f3f3f3} &\cellcolor[HTML]{f3f3f3} &\cellcolor[HTML]{ffecb3}--- &\cellcolor[HTML]{f3f3f3} &\cellcolor[HTML]{f3f3f3} &\cellcolor[HTML]{f3f3f3} &\cellcolor[HTML]{f3f3f3} &\cellcolor[HTML]{f3f3f3} &\cellcolor[HTML]{f3f3f3} &\cellcolor[HTML]{f3f3f3} &\cellcolor[HTML]{f3f3f3} &\cellcolor[HTML]{f3f3f3} &\cellcolor[HTML]{f3f3f3} &\cellcolor[HTML]{f3f3f3} &\cellcolor[HTML]{f3f3f3} &\cellcolor[HTML]{f3f3f3} &\cellcolor[HTML]{f3f3f3} &\cellcolor[HTML]{f3f3f3} &\cellcolor[HTML]{f3f3f3} &\cellcolor[HTML]{f3f3f3} &\cellcolor[HTML]{f3f3f3} &\cellcolor[HTML]{f3f3f3}
\\
&\textbf{Model Weight Randomization \citep{arun2020assessing}} &\cellcolor[HTML]{d9ead3}\cmark &\cellcolor[HTML]{d9ead3}\cmark &\cellcolor[HTML]{d9ead3}\cmark &\cellcolor[HTML]{d9ead3}\cmark &\cellcolor[HTML]{ffecb3}--- &\cellcolor[HTML]{ffecb3}--- &\cellcolor[HTML]{d9ead3}\cmark &\cellcolor[HTML]{f3f3f3} &\cellcolor[HTML]{f4cccc}\xmark &\cellcolor[HTML]{f3f3f3} &\cellcolor[HTML]{f3f3f3} &\cellcolor[HTML]{f3f3f3} &\cellcolor[HTML]{f3f3f3} &\cellcolor[HTML]{f3f3f3} &\cellcolor[HTML]{f3f3f3} &\cellcolor[HTML]{f3f3f3} &\cellcolor[HTML]{f3f3f3} &\cellcolor[HTML]{f3f3f3} &\cellcolor[HTML]{f3f3f3} &\cellcolor[HTML]{f3f3f3} &\cellcolor[HTML]{f3f3f3} &\cellcolor[HTML]{f3f3f3} &\cellcolor[HTML]{f3f3f3} &\cellcolor[HTML]{f3f3f3} &\cellcolor[HTML]{f3f3f3} 
\\
&\textbf{Repeatability \citep{arun2020assessing}} &\cellcolor[HTML]{f4cccc}\xmark &\cellcolor[HTML]{f4cccc}\xmark &\cellcolor[HTML]{ffecb3}--- &\cellcolor[HTML]{ffecb3}--- &\cellcolor[HTML]{ffecb3}--- &\cellcolor[HTML]{f4cccc}\xmark &\cellcolor[HTML]{f4cccc}\xmark &\cellcolor[HTML]{f3f3f3} &\cellcolor[HTML]{d9ead3}\cmark &\cellcolor[HTML]{f3f3f3} &\cellcolor[HTML]{f3f3f3} &\cellcolor[HTML]{f3f3f3} &\cellcolor[HTML]{f3f3f3} &\cellcolor[HTML]{f3f3f3} &\cellcolor[HTML]{f3f3f3} &\cellcolor[HTML]{f3f3f3} &\cellcolor[HTML]{f3f3f3} &\cellcolor[HTML]{f3f3f3} &\cellcolor[HTML]{f3f3f3} &\cellcolor[HTML]{f3f3f3} &\cellcolor[HTML]{f3f3f3} &\cellcolor[HTML]{f3f3f3} &\cellcolor[HTML]{f3f3f3} &\cellcolor[HTML]{f3f3f3} &\cellcolor[HTML]{f3f3f3} 
\\
&\textbf{Reproducibility \citep{arun2020assessing}} &\cellcolor[HTML]{f4cccc}\xmark &\cellcolor[HTML]{f4cccc}\xmark &\cellcolor[HTML]{f4cccc}\xmark &\cellcolor[HTML]{ffecb3}--- &\cellcolor[HTML]{ffecb3}--- &\cellcolor[HTML]{f4cccc}\xmark &\cellcolor[HTML]{f4cccc}\xmark &\cellcolor[HTML]{f3f3f3} &\cellcolor[HTML]{d9ead3}\cmark &\cellcolor[HTML]{f3f3f3} &\cellcolor[HTML]{f3f3f3} &\cellcolor[HTML]{f3f3f3} &\cellcolor[HTML]{f3f3f3} &\cellcolor[HTML]{f3f3f3} &\cellcolor[HTML]{f3f3f3} &\cellcolor[HTML]{f3f3f3} &\cellcolor[HTML]{f3f3f3} &\cellcolor[HTML]{f3f3f3} &\cellcolor[HTML]{f3f3f3} &\cellcolor[HTML]{f3f3f3} &\cellcolor[HTML]{f3f3f3} &\cellcolor[HTML]{f3f3f3} &\cellcolor[HTML]{f3f3f3} &\cellcolor[HTML]{f3f3f3} &\cellcolor[HTML]{f3f3f3} \\
\midrule
{\scriptsize\textcolor[HTML]{ffb000}{\textbf{Minimality}}} &\textbf{Minimality \citep{carter2019made}} &\cellcolor[HTML]{f3f3f3} &\cellcolor[HTML]{f3f3f3} &\cellcolor[HTML]{f3f3f3} &\cellcolor[HTML]{f3f3f3} &\cellcolor[HTML]{f3f3f3} &\cellcolor[HTML]{f3f3f3} &\cellcolor[HTML]{f3f3f3} &\cellcolor[HTML]{f3f3f3} &\cellcolor[HTML]{f3f3f3} &\cellcolor[HTML]{f3f3f3} &\cellcolor[HTML]{f3f3f3} &\cellcolor[HTML]{f3f3f3} &\cellcolor[HTML]{f3f3f3} &\cellcolor[HTML]{f3f3f3} &\cellcolor[HTML]{f3f3f3} &\cellcolor[HTML]{d9ead3}\cmark &\cellcolor[HTML]{f3f3f3} &\cellcolor[HTML]{f3f3f3} &\cellcolor[HTML]{f3f3f3} &\cellcolor[HTML]{f3f3f3} &\cellcolor[HTML]{f3f3f3} &\cellcolor[HTML]{f3f3f3} &\cellcolor[HTML]{f3f3f3} &\cellcolor[HTML]{f3f3f3} &\cellcolor[HTML]{f3f3f3} 
\\
&\textbf{Sparsity \citep{gomez2022metrics}} &\cellcolor[HTML]{f3f3f3} &\cellcolor[HTML]{f3f3f3} &\cellcolor[HTML]{f3f3f3} &\cellcolor[HTML]{ffecb3}--- &\cellcolor[HTML]{f3f3f3} &\cellcolor[HTML]{f3f3f3} &\cellcolor[HTML]{f3f3f3} &\cellcolor[HTML]{f3f3f3} &\cellcolor[HTML]{f3f3f3} &\cellcolor[HTML]{f3f3f3} &\cellcolor[HTML]{f3f3f3} &\cellcolor[HTML]{d9ead3}\cmark &\cellcolor[HTML]{ffecb3}--- &\cellcolor[HTML]{ffecb3}--- &\cellcolor[HTML]{ffecb3}--- &\cellcolor[HTML]{f3f3f3} &\cellcolor[HTML]{f3f3f3} &\cellcolor[HTML]{f3f3f3} &\cellcolor[HTML]{f3f3f3} &\cellcolor[HTML]{f3f3f3} &\cellcolor[HTML]{f3f3f3} &\cellcolor[HTML]{f3f3f3} &\cellcolor[HTML]{f3f3f3} &\cellcolor[HTML]{f3f3f3} &\cellcolor[HTML]{f3f3f3} \\
&\textbf{Visual Sharpening \citep{smilkov2017smoothgrad}} &\cellcolor[HTML]{f4cccc}\xmark &\cellcolor[HTML]{d9ead3}\cmark &\cellcolor[HTML]{ffecb3}--- &\cellcolor[HTML]{f3f3f3} &\cellcolor[HTML]{f3f3f3} &\cellcolor[HTML]{f4cccc}\xmark &\cellcolor[HTML]{d9ead3}\cmark &\cellcolor[HTML]{f3f3f3}     &\cellcolor[HTML]{f3f3f3} &\cellcolor[HTML]{f3f3f3} &\cellcolor[HTML]{f3f3f3} &\cellcolor[HTML]{f3f3f3} &\cellcolor[HTML]{f3f3f3} &\cellcolor[HTML]{f3f3f3} &\cellcolor[HTML]{f3f3f3} &\cellcolor[HTML]{f3f3f3} &\cellcolor[HTML]{f3f3f3} &\cellcolor[HTML]{f3f3f3} &\cellcolor[HTML]{f3f3f3} &\cellcolor[HTML]{f3f3f3} &\cellcolor[HTML]{f3f3f3} &\cellcolor[HTML]{d9ead3}\cmark &\cellcolor[HTML]{f3f3f3} &\cellcolor[HTML]{f3f3f3} &\cellcolor[HTML]{f3f3f3} \\
\midrule
{\scriptsize\textcolor[HTML]{ffb000}{\textbf{Perceptual}}} &\textbf{Localization Utility \citep{arun2020assessing}} &\cellcolor[HTML]{f4cccc}\xmark &\cellcolor[HTML]{f4cccc}\xmark &\cellcolor[HTML]{f4cccc}\xmark &\cellcolor[HTML]{f4cccc}\xmark &\cellcolor[HTML]{f4cccc}\xmark &\cellcolor[HTML]{f4cccc}\xmark &\cellcolor[HTML]{f4cccc}\xmark &\cellcolor[HTML]{f3f3f3} &\cellcolor[HTML]{d9ead3}\cmark &\cellcolor[HTML]{f3f3f3} &\cellcolor[HTML]{f3f3f3} &\cellcolor[HTML]{f3f3f3} &\cellcolor[HTML]{f3f3f3} &\cellcolor[HTML]{f3f3f3} &\cellcolor[HTML]{f3f3f3} &\cellcolor[HTML]{f3f3f3} &\cellcolor[HTML]{f3f3f3} &\cellcolor[HTML]{f3f3f3} &\cellcolor[HTML]{f3f3f3} &\cellcolor[HTML]{f3f3f3} &\cellcolor[HTML]{f3f3f3} &\cellcolor[HTML]{f3f3f3} &\cellcolor[HTML]{f3f3f3} &\cellcolor[HTML]{f3f3f3} &\cellcolor[HTML]{f3f3f3} \\
{\scriptsize\textcolor[HTML]{ffb000}{\textbf{Correspondence}}} &\textbf{Luminosity Calibration \citep{gomez2022metrics}} &\cellcolor[HTML]{f3f3f3} &\cellcolor[HTML]{f3f3f3} &\cellcolor[HTML]{f3f3f3} &\cellcolor[HTML]{f4cccc}\xmark &\cellcolor[HTML]{f3f3f3} &\cellcolor[HTML]{f3f3f3} &\cellcolor[HTML]{f3f3f3} &\cellcolor[HTML]{f3f3f3} &\cellcolor[HTML]{f3f3f3} &\cellcolor[HTML]{f3f3f3} &\cellcolor[HTML]{f3f3f3} &\cellcolor[HTML]{f4cccc}\xmark &\cellcolor[HTML]{f4cccc}\xmark &\cellcolor[HTML]{d9ead3}\cmark &\cellcolor[HTML]{f4cccc}\xmark &\cellcolor[HTML]{f3f3f3} &\cellcolor[HTML]{f3f3f3} &\cellcolor[HTML]{f3f3f3} &\cellcolor[HTML]{f3f3f3} &\cellcolor[HTML]{f3f3f3} &\cellcolor[HTML]{f3f3f3} &\cellcolor[HTML]{f3f3f3} &\cellcolor[HTML]{f3f3f3} &\cellcolor[HTML]{f3f3f3} &\cellcolor[HTML]{f3f3f3} \\
&\textbf{Mean IoU \citep{saporta2022benchmarking}} &\cellcolor[HTML]{f3f3f3} &\cellcolor[HTML]{f3f3f3} &\cellcolor[HTML]{f3f3f3} &\cellcolor[HTML]{ffecb3}---  &\cellcolor[HTML]{ffecb3}--- &\cellcolor[HTML]{f4cccc}\xmark &\cellcolor[HTML]{f3f3f3} &\cellcolor[HTML]{f3f3f3} &\cellcolor[HTML]{f3f3f3} &\cellcolor[HTML]{f4cccc}\xmark &\cellcolor[HTML]{f4cccc}\xmark &\cellcolor[HTML]{f3f3f3} &\cellcolor[HTML]{f3f3f3} &\cellcolor[HTML]{f3f3f3} &\cellcolor[HTML]{f3f3f3} &\cellcolor[HTML]{f3f3f3} &\cellcolor[HTML]{f3f3f3} &\cellcolor[HTML]{f3f3f3} &\cellcolor[HTML]{ffecb3}--- &\cellcolor[HTML]{ffecb3}--- &\cellcolor[HTML]{f3f3f3} &\cellcolor[HTML]{f3f3f3} &\cellcolor[HTML]{f3f3f3} &\cellcolor[HTML]{f3f3f3} &\cellcolor[HTML]{f3f3f3} \\
&\textbf{Plausibility \citep{ding2021evaluating}} &\cellcolor[HTML]{ffecb3}--- &\cellcolor[HTML]{d9ead3}\cmark &\cellcolor[HTML]{f3f3f3} &\cellcolor[HTML]{f3f3f3} &\cellcolor[HTML]{f3f3f3} &\cellcolor[HTML]{d9ead3}\cmark &\cellcolor[HTML]{f3f3f3} &\cellcolor[HTML]{f3f3f3} &\cellcolor[HTML]{f3f3f3} &\cellcolor[HTML]{f3f3f3} &\cellcolor[HTML]{f3f3f3} &\cellcolor[HTML]{f3f3f3} &\cellcolor[HTML]{f3f3f3} &\cellcolor[HTML]{f3f3f3} &\cellcolor[HTML]{f3f3f3} &\cellcolor[HTML]{f3f3f3} &\cellcolor[HTML]{f3f3f3} &\cellcolor[HTML]{f3f3f3} &\cellcolor[HTML]{f3f3f3} &\cellcolor[HTML]{f3f3f3} &\cellcolor[HTML]{f3f3f3} &\cellcolor[HTML]{f3f3f3} &\cellcolor[HTML]{f3f3f3} &\cellcolor[HTML]{f3f3f3} &\cellcolor[HTML]{f3f3f3} \\
&\textbf{The Pointing Game \citep{zhang2016top} / Hit Rate \citep{saporta2022benchmarking}} &\cellcolor[HTML]{d9ead3}\cmark &\cellcolor[HTML]{f3f3f3} &\cellcolor[HTML]{f3f3f3} &\cellcolor[HTML]{ffecb3}--- &\cellcolor[HTML]{f3f3f3} &\cellcolor[HTML]{ffecb3}--- &\cellcolor[HTML]{f3f3f3} &\cellcolor[HTML]{f3f3f3} &\cellcolor[HTML]{f3f3f3} &\cellcolor[HTML]{ffecb3}--- &\cellcolor[HTML]{ffecb3}--- &\cellcolor[HTML]{f3f3f3} &\cellcolor[HTML]{f3f3f3} &\cellcolor[HTML]{f3f3f3} &\cellcolor[HTML]{ffecb3}--- &\cellcolor[HTML]{f3f3f3} &\cellcolor[HTML]{d9ead3}\cmark &\cellcolor[HTML]{d9ead3}\cmark &\cellcolor[HTML]{f4cccc}\xmark &\cellcolor[HTML]{ffecb3}--- &\cellcolor[HTML]{f3f3f3} &\cellcolor[HTML]{f3f3f3} &\cellcolor[HTML]{f3f3f3} &\cellcolor[HTML]{f3f3f3} &\cellcolor[HTML]{f3f3f3}  \\
\bottomrule
\end{tabular}
}
\end{table*}
\clearpage

\subsection{Additional Interview Details}
\label{sup:user-study}
\begin{figure*}[ht]
  \centering
  \includegraphics[width=0.87\textwidth]{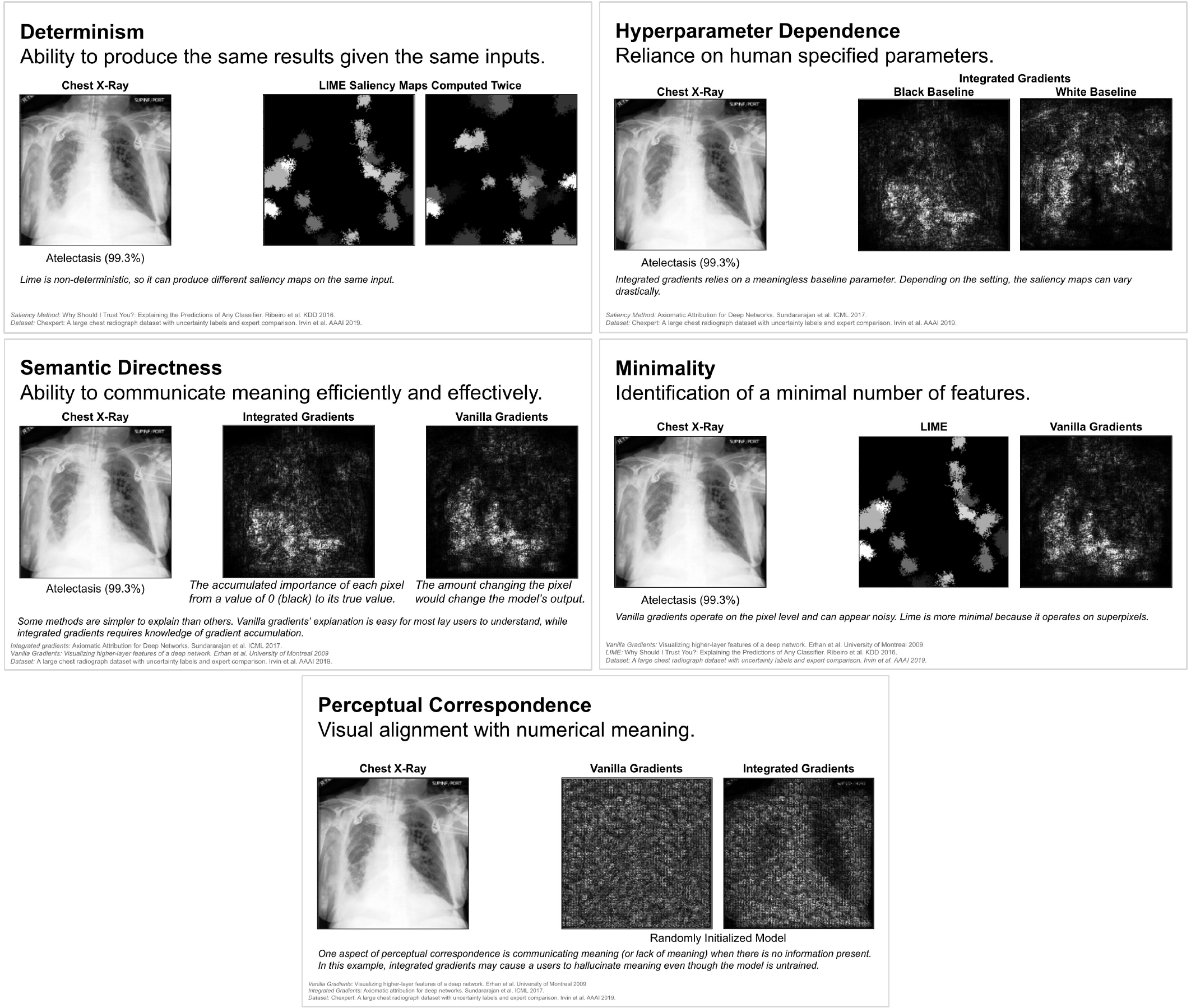}
  \caption{Examples of each saliency card attribute we discussed with the radiologist user (\texttt{U8}) in our user study. Given they were unfamiliar with machine learning, we used medical imaging examples from CheXpert~\citep{irvin2019chexpert}. Each example defines the attribute and shows an informative instance of the attribute exhibited.}
  \label{fig:radiology-user-study-slides}
  \Description{Five slides show examples of determinism, hyperparameter dependence, semantic directness, minimality, and perceptual correspondence using a chest x-ray image.}
\end{figure*}

\begin{figure*}[ht]
  \centering
  \includegraphics[width=0.82\textwidth]{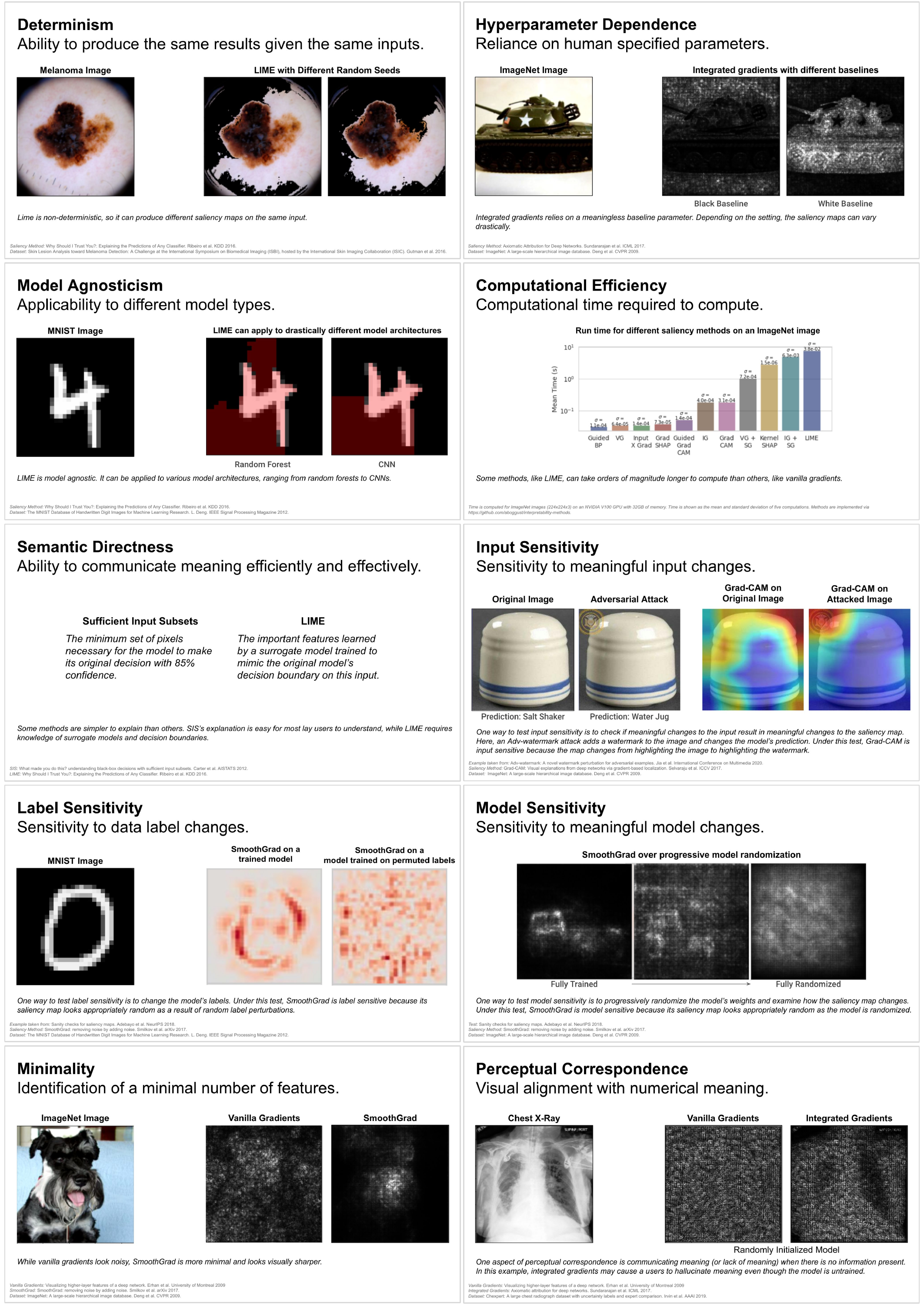}
  \caption{Examples of each saliency card attribute we discussed with the \texttt{U1}--\texttt{U7}. Each example defines the attribute and shows an informative instance of the attribute exhibited.}
  \label{fig:user-study-slides}
  \Description{Ten slides show example of each of the ten saliency card attributes on computer vision tasks.}
\end{figure*}

In each interview, we explained saliency card attributes via a definition and example that demonstrated the attribute.
We showed \texttt{U1}--\texttt{U8} examples from ImageNet~\citep{deng2009imagenet}, melanoma classification~\citep{codella2018skin}, MNIST digit recognition~\citep{deng2012mnist}, and CheXpert chest x-rays~\citep{irvin2019chexpert}, shown in Fig.~\ref{fig:user-study-slides}.
Since the radiologist participant (\texttt{U9}) was unfamiliar with machine learning, we only showed them examples using CheXpert chest x-rays~\citep{irvin2019chexpert}, shown in Fig.~\ref{fig:radiology-user-study-slides}.

\end{document}